%% file: graphac-iclr2023-mldd.tex
\PassOptionsToPackage{table}{xcolor}

\documentclass{article} 
\usepackage{MLDD_workshop_2023,times}

\usepackage{booktabs} 
\usepackage{graphicx}
\usepackage{makecell}
\usepackage{microtype}
\usepackage{multirow}
\usepackage{subcaption}
\usepackage{xcolor} 
\usepackage[export]{adjustbox}

\newcolumntype{x}[1]{>{\centering\arraybackslash\hspace{0pt}}p{#1}}

\usepackage{hyperref}
\usepackage{url}

\usepackage{algorithm}
\usepackage{algorithmic}
\usepackage{amsmath}
\usepackage{amssymb}
\usepackage{mathtools}
\usepackage{amsthm}
\usepackage{nicefrac}
\usepackage{listings}

\lstdefinelanguage{myPython}{
	language=Python,
	basicstyle=\ttfamily\small,
	commentstyle=\color{gray},
	morekeywords=[1]{as, assert, with, True, False},
	deletekeywords=[2]{sum, pow},
	upquote=true,
	escapeinside={``},
	columns=flexible,
	keepspaces=true,
	numbers=none,
	showstringspaces=false,
}

\title{Task-Agnostic Graph Neural Network \\ Evaluation via Adversarial Collaboration}

\author{Xiangyu Zhao\textsuperscript{1}, Hannes St{\"a}rk\textsuperscript{2}, Dominique Beaini\textsuperscript{3}, Yiren Zhao\textsuperscript{1}, Pietro Li{\`o}\textsuperscript{4} \\
\textsuperscript{1}Department of Electrical and Electronic Engineering, Imperial College London \\
\textsuperscript{2}CSAIL, Massachusetts Institute of Technology \\
\textsuperscript{3}Valence Discovery, Mila, Universit{\'e} de Montr{\'e}al \\
\textsuperscript{4}Department of Computer Science and Technology, University of Cambridge \\
\textsuperscript{1}\texttt{\{x.zhao22,a.zhao\}@imperial.ac.uk}, \\
\textsuperscript{2}\texttt{hstark@mit.edu}, 
\textsuperscript{4}\texttt{pietro.lio@cl.cam.ac.uk}
}

\iclrfinalcopy 
\begin{document}

\maketitle

\begin{abstract}
It has been increasingly demanding to develop reliable methods to evaluate the progress of Graph Neural Network (GNN) research for molecular representation learning. Existing GNN benchmarking methods for molecular representation learning focus on comparing the GNNs' performances on some node/graph classification/regression tasks on certain datasets. However, there lacks a principled, task-agnostic method to directly compare two GNNs. Additionally, most of the existing self-supervised learning works incorporate handcrafted augmentations to the data, which has several severe difficulties to be applied on graphs due to their unique characteristics. To address the aforementioned issues, we propose GraphAC (Graph Adversarial Collaboration) -- a conceptually novel, principled, task-agnostic, and stable framework for evaluating GNNs through contrastive self-supervision. We introduce a novel objective function: the Competitive Barlow Twins, that allow two GNNs to jointly update themselves from direct competitions against each other. GraphAC succeeds in distinguishing GNNs of different expressiveness across various aspects, and has demonstrated to be a principled and reliable GNN evaluation method, without necessitating any augmentations.
\end{abstract}

\input{sections/introduction.tex}
\input{sections/related-work.tex}
\input{sections/method.tex}
\input{sections/evaluation.tex}

\input{sections/conclusion.tex}
\input{sections/acknowledgements.tex}

\bibliography{references}
\bibliographystyle{iclr2023_conference}

\newpage
\appendix
\input{appendices/algorithm.tex}
\input{appendices/results-details.tex}
\input{appendices/correlation-study.tex}
\input{appendices/training-details.tex}

\end{document}

%% file: sections/introduction.tex
\section{Introduction}

Graph Neural Networks (GNNs) have gained immense research attention in recent years, leading to significant progress that has been successfully implemented across a broad range of fields, including chemistry \citep{Gilmer2017MPNN} and biology \citep{Stokes2020Antibody}. This makes GNNs important tools in the molecular representation learning landscape, and improving their development is of great interest to the biomedical machine learning community. 
As this field of research grows rapidly, it becomes crucial to develop general and reliable evaluation methods to facilitate GNN research and quantify the performances of various GNN architectures in the context of molecular graph data. 

Existing approaches on benchmarking GNNs focus on comparing the GNNs with respect to their performances on some node/graph classification/regression tasks in a collection of molecular/protein/DNA datasets \citep{Dwivedi2020Benchmark,Hu2020OGB}. However, these approaches can be limited in the following ways: 1) classification/regression tasks are naturally simple in terms of the combinatorial complexity, and cannot fully challenge the GNNs to learn from the graphs; 2) there exist many different molecular prediction datasets used by various works on GNNs, but there lacks a standardized way to compare those results on different datasets, and sometimes it is not feasible to evaluate on all of them; and 3) Deep Neural Networks (DNNs) generally rely on high-quality labeled data for high-performance training, but unfortunately, large datasets almost always contain examples with inaccurate, incorrect or even missing labels, known as label noises. It has been demonstrated that most DNNs, especially GNNs, are vulnerable to noisy labels, resulting in drastically lowered generalization performances (\citealp{Dai2021NRGNN,NT2019GNNNoisy,Zhang2017Understanding}; \citealp{Zhang2020Adversarial}). Consequently, it is highly desirable to develop GNN evaluation methods that can effectively exploit the training data, without necessitating any labels.

There have been some attempts to evaluate the capacities of GNNs against theoretical tests such as the Weisfeiler-Lehman (1-WL) graph isomorphism test \citep{Weisfeiler1968WL,Xu2019GIN,Dwivedi2020Benchmark}, but the information they provide is normally limited. These methods are normally designed for small-scale benchmarks, and measure GNNs' ability to detect patterns, substructures and clusters \citep{Dwivedi2020Benchmark}, or graph properties such as diameter, eccentricity, and spectral radius \citep{Corso2020PNA}. However, these approaches cannot be used consistently, since certain GNN types can exploit this information as positional encodings to directly cheat the task \citep{Kreuzer2021Rethinking,Bodnar2021CWN,Dwivedi2022LSPE}. Therefore, there is a need for a task-agnostic evaluation of the expressiveness of GNNs.

Designing principled self-supervised learning (SSL) methods for graphs is also a challenging task. As labeled data can be expensive, limited or even unavailable in many real-word scenarios, it has become increasingly demanding to develop powerful SSL methods on graphs. A lot of successful SSL works on graphs \citep{Velickovic2019DGI,Sun2020InfoGraph,You2020GraphCL,You2021GraphCLAutomated,Xu2021GraphLoG} rely on applying handcrafted augmentations to the graphs, which are further described in Section~\ref{sec:related-work}. However, there are several key difficulties in applying augmentations to graphs. Firstly, there exists no universal augmentation that works across all types of graphs. Secondly, graphs are not invariant to augmentations like images: applying filters or rotating an image still preserves its essential invariances, but even a tiny augmentation on a graph can significantly change its topological structure or intrinsic properties. Another class of SSL methods on graphs that do not require augmentations \citep{Stark20213DInfomax} relies on exploiting the physical properties of small molecules, and cannot be generalized to other graph types, such as proteins or DNAs. Therefore, it is highly desirable to develop a principled and generalizable SSL framework that does not require handcrafted augmentations. 

\textbf{Our solution: Graph Adversarial Collaboration (GraphAC).} We address both aforementioned questions by proposing a conceptually novel, principled, and task-agnostic framework for evaluating GNNs in the context of molecular data, via a self-supervised, adversarial collaboration manner, without the need of handcrafted augmentations. In the GraphAC framework, two GNNs directly compete against each other on the same unlabeled graphs. The more expressive GNN produces more complex and informative graph embeddings and is thereby able to win the game. We make the following contributions in this paper:

\begin{itemize}
    \item We introduce a novel principle for evaluating GNNs, by having them directly compete against each other in a self-supervised manner, rather than comparing them using a scoreboard of training performances on some datasets;
    \item Inspired by the novel principle, we propose a new architecture and an original modification to the existing Barlow Twins loss \citep{Zbontar2021BarlowTwins} that enables the GNNs to stably compete against each other, while ensuring that more expressive GNNs can always win;
    \item We provide the very first framework for evaluating GNN expressiveness directly on the molecular graph data, without the need of a specific downstream task, or theoretical representations of these molecular graphs;
    \item We develop a principled contrastive learning framework without needing any handcrafted augmentations, which is also generalizable to various types of GNNs.
\end{itemize}

%% file: sections/related-work.tex
\section{Related Work} \label{sec:related-work}



\paragraph{Contrastive Self-Supervised Learning} To the best of our knowledge, there has not been any published attempt to develop a method for evaluating deep learning models by directly competing two models in a contrastive self-supervised environment, no matter in the general machine learning or the graph representation learning communities. Current approaches center around competing with a set of baselines that evaluate a limited number of performance metrics on a fixed number of benchmark or datasets. However, the state-of-the-arts in contrastive SSL, both in the non-graph and the graph domains, is still relevant to this work. Their successes in building contrastive learning architectures can help us build a principled, task-agnostic graph model evaluation framework. It is worth noting that GraphAC is a task-agnostic evaluation of GNNs, and not an optimization for downstream tasks. Therefore, the performance of state-of-the-art contrastive SSL on downstream tasks is not relevant.

\newpage
Many works on contrastive SSL on graphs \citep{Velickovic2019DGI,Sun2020InfoGraph,You2020GraphCL,You2021GraphCLAutomated,Xu2021GraphLoG} are inspired from the successes of the idea of mutual information maximization between two representations of the same data, with manually applied augmentations, in the non-graph domain \citep{Gutmann2010NCE,Oord2018InfoNCE,Hjelm2019DIM,He2020MoCo,Chen2020SimCLR}. Those works vary in augmentation strategies and mutual information estimators. However, in order to prevent \emph{information collapse} (models ignoring the input data and outputting identical and constant vectors), all those works require large batch sizes or memory banks, and extensive searches for augmentations and negative pairs, making them very costly. Besides, applying augmentations to graphs can be much harder than to images, since there exists no universal augmentation that works for all graph types. Furthermore, graphs are not noise-invariant -- small changes to a graph can significantly alter its topological structure, especially for small graphs such as molecules. While existing research has developed fine-grained graph augmentations, it is still almost impossible to apply these augmentations while preserving the graph's intrinsic properties, such as the chemical properties of molecules. Moreover, graph augmentations can deviate the data from real-word distributions, since they introduce arbitrary human knowledge not provided by the training data. \citet{Stark20213DInfomax} propose a noiseless framework by maximizing the mutual information between the embedding of a 2D molecular graph and the embedding capturing its 3D structure, but it is specific to the physical properties of molecules, and cannot be generalized to other domains. Consequently, \emph{there is a need for a principled contrastive SSL framework that can be applied across a diversity of graph types without requiring any augmentations}.

\paragraph{Barlow Twins} \citet{Zbontar2021BarlowTwins} introduce an alternative approach to prevent information collapse, by maximizing the information contents within the representations. In Barlow Twins, for a given input batch $\mathbf{X}\in\mathbb{R}^{N_b\times d_\mathbf{x}}$ of batch size $N_b$ and dimension $d_\mathbf{x}$, two batches of distorted views $\tilde{\mathbf{X}}^A$ and $\tilde{\mathbf{X}}^B$ of $\mathbf{X}$ are generated using manual data augmentation. The two batches of distorted views $\tilde{\mathbf{X}}_A$ and $\tilde{\mathbf{X}}^B$ are fed into two separate models, which produces batches of $d$-dimensional embeddings $\mathbf{H}^A, \mathbf{H}^B\in\mathbb{R}^{N_b\times d}$. For simplicity,  the features in both $\mathbf{H}^A$ and $\mathbf{H}^B$ are assumed to have a mean of zero across the batch. Barlow Twins then computes the cross-correlation $\mathbf{C}\in\mathbb{R}^{d\times d}$ matrix between $\mathbf{H}^A$ and $\mathbf{H}^B$ along the batch dimension. It then applies the following loss function on $\mathbf{C}$: 

\begin{equation}
	\mathcal{L}_\text{BT}=\underbrace{\sum_i^d(1-\mathbf{C}_{i,i})^2}_{\text{invariance term}}+\lambda\underbrace{\sum_i^d\sum_{j\neq i}^d \mathbf{C}_{i,j}^2}_{\text{redundancy reduction term}}
\end{equation}

The invariance term of the Barlow Twins loss enforces the two output embeddings to be similar by pushing the on-diagonal elements of the cross-correlation matrix towards one. Meanwhile, the redundancy reduction term attempts to ensure that the off-diagonal elements of the cross-correlation matrix closer to zero, thereby decorrelating the different features of the embeddings, so that the embeddings contain non-redundant information about the data. This process implicitly maximizes the amount of information contained within the embedding vectors.

\paragraph{Variance-Invariance-Covariance Regularization (VICReg)} \citet{Bardes2022VICReg} build VICReg based on the principle of preserving the information content of the representations, similar to Barlow Twins. The architecture of VICReg is the same as Barlow Twins, except that it uses three regularization terms in its objective function: 1) invariance regularization $\mathcal{L}_\text{Inv}$: the mean square Euclidean distance between the output embeddings; 2) variance regularization $\mathcal{L}_\text{Var}$: a hinge loss to maintain the standard deviation of the embeddings along the batch dimension close to 1, which forces the output embeddings within a batch to be different; and 3) covariance regularization $\mathcal{L}_\text{Cov}$: the sum of the squared off-diagonal elements of the covariance matrix, with a factor $\nicefrac{1}{d}$ to scale the term as a function of the feature dimension. This term attracts the covariances between every pair of features of the embeddings over a batch towards zero, decorrelating the different features of the embeddings, thus preventing them from encoding similar information.
The overall loss function for VICReg is then a weighted sum of the invariance, variance and covariance regularization terms:

\begin{equation}
    \mathcal{L}_\text{VICReg} = \lambda\mathcal{L}_\text{Inv}+\mu\mathcal{L}_\text{Var}+\nu\mathcal{L}_\text{Cov}
\end{equation}

where $\lambda,\mu,\nu>0$ are hyperparameters controlling the importance of each term in the loss.

%% file: sections/method.tex
\section{Method} \label{sec:method}

The intuition behind Graph Adversarial Collaboration (GraphAC) is to have different GNNs competing against each other on the same \emph{unlabeled} graphs, and encouraging more expressive GNNs to produce more complex and informative graph embeddings. This can be measured by the ability to predict other GNNs' graph embeddings from a GNN's own graph embeddings: if a GNN can predict another GNN's graph embeddings from its own graph embeddings better than the other way round, then its graph embeddings can be deemed more complex and informative than the other GNN's graph embeddings, and therefore, more expressive. The two GNNs collaborate by predicting each other's output graph embeddings, and compete adversarially to prevent the other GNNs from predicting their own graph embeddings. To solve the challenge of maximizing the performance differences between different GNNs while ensuring stable training, we introduce the \emph{Competitive Barlow Twins}, a novel pair of loss functions modified from the Barlow Twins described in Section~\ref{sec:related-work}.

\subsection{Competitive Barlow Twins} \label{sec:competitive-bt}

A deeper analysis of the Barlow Twins shows that, according to \citet{Zbontar2021BarlowTwins}'s definition for the cross-correlation matrix $\mathbf{C}$ between the embeddings, 

\begin{equation}
    \mathbf{C}_{i,j}=\frac{\sum_b^{N_b}\mathbf{H}^A_{b,i}\mathbf{H}^B_{b,j}}{\sqrt{\sum_b^{N_b}\big(\mathbf{H}^A_{b,i}\big)^2}\sqrt{\sum_b^{N_b}\big(\mathbf{H}^B_{b,j}\big)^2}} 
\end{equation}

the $(i,j)$-th entry $\mathbf{C}_{i,j}$ of the cross-correlation matrix represents how much feature $i$ of the first model's output embeddings $\mathbf{H}^A$ correlates to feature $j$ of the second model's output embeddings $\mathbf{H}^B$. Therefore, for output embeddings of dimensionality $d$, the row $\mathbf{C}_{i,[i+1:d]}$ at the upper-triangle of the cross-correlation matrix represents how much feature $i$ of $\mathbf{H}^A$ correlates to features $i+1$ to $d$ of $\mathbf{H}^B$. For $i$ close to one, the row $\mathbf{C}_{i,[i+1:d]}$ in the upper-triangle becomes much longer, and thus the $i$-th feature of $\mathbf{H}^A$ represented by that piece of the row correlates to the majority of the features of $\mathbf{H}^B$. For $i$ close to $d$, the row at the upper-triangle becomes much shorter, and thus the $i$-th feature of $\mathbf{H}^A$ represented by that piece of the row correlates to very few features of $\mathbf{H}^B$. This means that the smaller-indexed features of the first model's output embeddings $\mathbf{H}^A$ become the more important features, if monitored by the upper-triangle of the cross-correlation matrix. Similarly, in the lower-triangle of the cross-correlation matrix, the column $\mathbf{C}_{[j+1:d],j}$ represents how much feature $j$ of $\mathbf{H}^B$ correlates to features $j+1$ to $d$ of $\mathbf{H}^A$, making the smaller-indexed features of the second model's output embeddings also becoming the more important features. It can therefore be hypothesized that under this upper-lower-triangle setting, the first few features of both models' output embeddings are targeted at capturing the low frequency signals as they are easier to predict, and the later features are set to capture the high frequency signals, which are harder to predict.

Based on the above findings, if the two triangles of the cross-correlation matrix are summed, then the sum of each triangle is dominated by the first few rows/columns, as they contain the most entries. Therefore, the sum of the triangle provides a measure of how much a model's output features correlate to the other model's output features, weighted by importance, since there are more elements in the triangle corresponding to the more important features. Consequently, a larger sum implies a better correlation of a model's most important features in its output embeddings with the other model's output features, which implies a stronger ability to predict the other model's output embeddings from its own output embeddings. This naturally yields the definition of the Competitive Barlow Twins loss, which preserves the invariance term in the original Barlow Twins, but replaces the off-diagonal sum with the difference between the upper-triangle and the lower-triangle of the cross-correlation matrix:

\begin{equation} \label{eqn:competitive-bt-loss}
	\begin{aligned}
		\mathcal{L}_{\text{CBT}_A} &= \sum_i^d(1-\mathbf{C}_{i,i})^2+\lambda\!\left(\!\sum_i^d\sum_{j>i}^d \mathbf{C}_{i,j}^2-\mu\!\sum_j^d\sum_{i>j}^d \mathbf{C}_{i,j}^2\!\right) \\
		\mathcal{L}_{\text{CBT}_B} &= \sum_i^d(1-\mathbf{C}_{i,i})^2+\lambda\!\left(\!\sum_j^d\sum_{i>j}^d \mathbf{C}_{i,j}^2-\mu\!\sum_i^d\sum_{j>i}^d \mathbf{C}_{i,j}^2\!\right) \\
	\end{aligned}
\end{equation}

where $\lambda,\mu>0$ are weighting coefficients, with $\lambda$ inherited from the original Barlow Twins, and $\mu$ trading off the importance of correlating the opponent GNN's output features (collaboration) and preventing the opponent GNN from correlating the GNN's own output features (competition). Although the above reasoning discourages the use of different weights on the sums of triangles, which is also confirmed by the hyperparameter tuning results described in Appendix~\ref{sec:hyperparam-tuning}, we still include $\mu$ in the definition of the Competitive Barlow Twins for the purpose of hyperparameter tuning.

Another important enhancement by the Competitive Barlow Twins is that, since the triangles make the smaller-indexed features of both models' output embeddings the more important features, both models' output embeddings are ordered by feature importance. This ordering prevents the models from simply permuting the entries of their output embeddings to avoid being predicted by their opponent models, and makes the training much more stable.

\begin{figure*}[t]
    \centering
    \includegraphics[width=\textwidth]{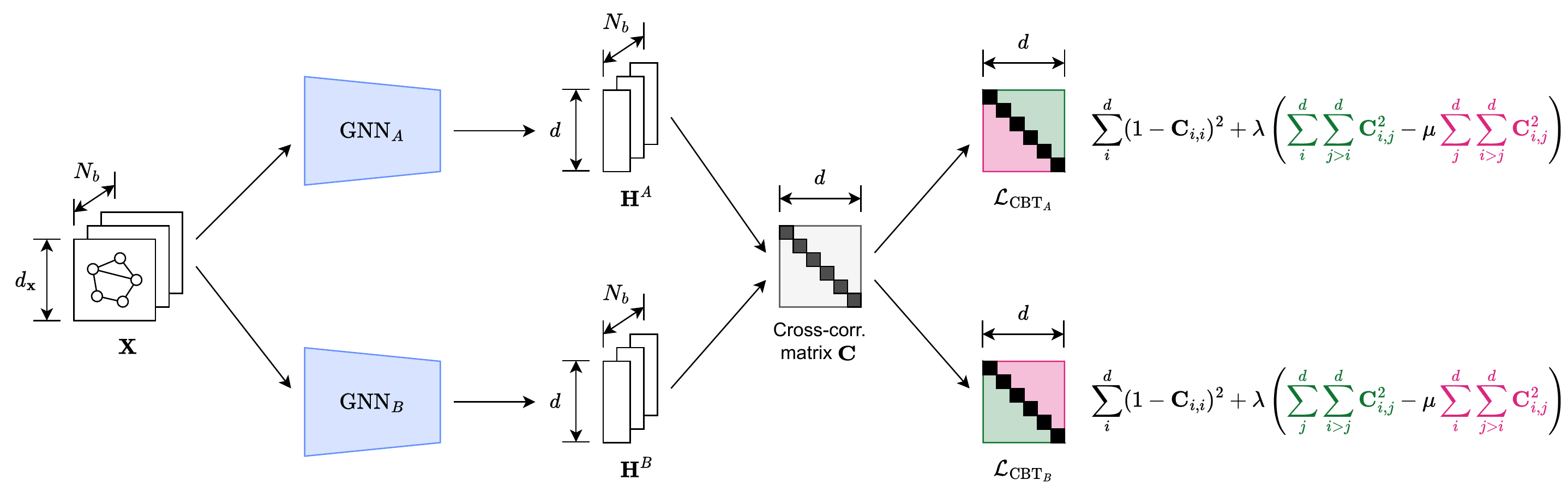}
    \caption{Architecture of GraphAC's framework. Batched unlabeled graph data $\mathbf{X}\in\mathbb{R}^{N_b\times d_\mathbf{x}}$ are fed into two different GNNs, obtaining two different batched embeddings $\mathbf{H}^A,\mathbf{H}^B\in\mathbb{R}^{N_b\times d}$. Then, the cross-correlation matrix $\mathbf{C}\in\mathbb{R}^{d\times d}$ between $\mathbf{H}^A$ and $\mathbf{H}^B$ along the batch dimension is calculated. The Competitive Barlow Twins losses, $\mathcal{L}_{\text{CBT}_A}$ and $\mathcal{L}_{\text{CBT}_B}$, are then computed by pushing the on-diagonal elements of $\mathbf{C}$ towards one, while adding the difference between the upper/lower-triangles of $\mathbf{C}$. The pair of Competitive Barlow Twins losses are then used to update the two GNNs respectively. Details about the GraphAC framework are described in Section~\ref{sec:method}.}
    \label{fig:competitive-bt}
\end{figure*}

\subsection{Proposed Framework}

The architecture of the GraphAC framework is illustrated in Figure~\ref{fig:competitive-bt}. We also include the covariance regularization term from VICReg in GraphAC's loss functions because it decorrelates different feature dimensions within each output graph embeddings, and forces the graph embeddings to be fully used to capture graph information. Therefore, we obtain the following definitions for GraphAC's final loss functions:

\begin{equation} \label{eqn:competitive-bt-cov-loss}
	\begin{aligned}
		\mathcal{L}_{\text{GNN}_A} &= \alpha\mathcal{L}_{\text{CBT}_A}+\beta\mathcal{L}_\text{Cov} \\
		\mathcal{L}_{\text{GNN}_B} &= \alpha\mathcal{L}_{\text{CBT}_B}+\beta\mathcal{L}_\text{Cov}
	\end{aligned}
\end{equation}

where $\alpha,\beta>0$ are weighting coefficients, and $\mathcal{L}_\text{Cov}$ is the VICReg covariance regularization term defined in Section~\ref{sec:related-work}. The invariance regularization term from VICReg is not included in the loss functions, because the effect of the invariance regularization term has already been achieved by the invariance term from the Competitive Barlow Twins. We also do not include the variance regularization term from VICReg in GraphAC's loss functions, because it forces the variance of the embeddings over a batch to be above a given threshold, which can potentially cause the training to be unstable. Although the Competitive Barlow Twins losses can enable stable training of the models, and can counter the instability caused by the VICReg variance regularization term, we still do not include that term in the loss functions, because it is used to prevent the models from producing the same embedding vectors for samples within a batch, which did not occur in this framework.

%% file: sections/evaluation.tex
\section{Evaluation}

\subsection{Data Preparation}

In order for GraphAC to provide the most realistic evaluations of the GNNs, the datasets used for evaluating it should be application-oriented with real-word implications. To ensure GraphAC can discriminate between GNNs with statistical significance, the datasets should be large-scale of high quality. Moreoever, since GraphAC is designed for graph-level prediction, the datasets should also be constructed for graph-level prediction, which means that they should contain a large number of relatively small graphs. Finally, in order for GraphAC to support GNNs both with and without edge features, and allowing it to study the effect for a GNN with edge features, the datasets should provide both node and edge features for the graphs. Based on the above requirements, GraphAC is best suited to drug-like small molecular datasets for the following reasons:

\begin{itemize}
    \item Molecules can naturally be represented as graphs;
    \item Molecular property prediction is a fundamental task within many important applications in chemistry, biology and medicine;
    \item There is a vast variety of molecules in the world, and the drug-like small molecular graphs can be trained efficiently without requiring extensive GPU resources.
\end{itemize}

Therefore, we use the largest molecular property prediction dataset from OGB \citep{Hu2020OGB}, namely the \texttt{ogbg-molpcba} dataset. It contains 437,929 drug-like molecules, with on average 26.0 atoms (nodes) and 28.1 bonds (edges) per molecule (graph). Despite the fact that the dataset comprises multiple classification tasks and its class balance is highly skewed, these factors do not affect GraphAC's evaluation, as only the molecular graphs are used while their labels are discarded. In order to confirm that GraphAC is indeed task-agnostic, we also evaluate GraphAC on the \texttt{ogbg-code2} dataset, which contains 452,741 abstract syntax trees obtained from Python method definitions, with on average 125.2 nodes and 124.2 edges per tree.

\subsection{Experimental Setup}

Training and experiments were conducted on an NVIDIA A100 SXM GPU with 80GB graphics memory. All experiments were trained for 50 epochs. The pseudocode for the core training algorithm of GraphAC can be found in Appendix~\ref{sec:pseudocode}, and details of the hyperparameter tuning experiments are described in Appendix~\ref{sec:hyperparam-tuning}. The source code for GraphAC is publicly available at \url{https://github.com/VictorZXY/GraphAC}.

In order to fairly compare the GNNs as well as to evaluate GraphAC's ability in distinguishing different components of a GNN, the experiments were split into five groups of controlled experiments. In each group, one component of the GNN is varied, while all other components are fixed. The five aspects of a GNN evaluated by GraphAC are:

\begin{itemize}
    \item \textbf{Number of GNN layers:} in this group of experiments, PNAs \citep{Corso2020PNA} with 2, 4, 6, 8, and 10 layers compete in the GraphAC framework on a double round-robin basis, with one extra experiment performed for each model to compete against itself. All PNAs have a fixed hidden dimension of 256, and use the combination of [max, mean, sum] as their aggregators. All PNAs use [identity, amplification, attenuation] as their scalers, and their message passing functions are parametrized by 2-layer MLPs. We choose PNAs for evaluation due to their flexibility and state-of-the-art performance on molecular tasks.
    \item \textbf{Hidden dimensions:} in this group of experiments, 4-layer PNAs with hidden dimenions of 16, 32, 64, 128, and 256 competed in the GraphAC framework on a double round-robin basis, with an additional experiment performed for each model to compete against itself. All PNAs utilize [max, mean, sum] as their aggregators.
    \item \textbf{Aggregators:} in this group of experiments, four 4-layer PNAs with 64 hidden dimensions, and [max], [mean], [sum], [max, mean, sum] as their aggregators respectively, are set to compete in the GraphAC framework on a double round-robin basis, again with one extra experiment for each model to compete against itself.
    \item \textbf{GNN architectures:} in this group of architectures, PNA, GIN \citep{Xu2019GIN} and GCN \citep{Kipf2017GCN}, all with 4 layers and 64 hidden dimensions, and PNA with [max, mean, sum] as aggregators, are set to compete in the GraphAC framework on a double round-robin basis, again with one extra experiment for each model to compete against itself.
    \item \textbf{Edge features:} in this group of experiments, PNAs with 4, 6, and 8 layers, and hidden dimensions ranging from 64, 128, and 256 are used. All PNAs use [max, mean, sum] as their aggregators. In each experiment, PNAs with the same structure, but one with the edge features and the other without the edge features, are inserted into GraphAC for competitions.
\end{itemize}

\subsection{Results}

\begin{table*}[t]\centering
    \caption{Loss differences of the conducted experiments. Negative value means $\text{GNN}_A$ wins the game, and positive value means $\text{GNN}_B$ wins the game. Greater absolute value indicates larger gap in expressiveness determined by GraphAC. The consistent gradient from bottom left to upper right clearly indicates that GraphAC genuinely favors more expressive GNNs. Since the \texttt{ogbg-code2} dataset does not contain any edge features, no experiments regarding the inclusion of edge features for the \texttt{ogbg-code2} dataset are conducted.}
    \label{tab:results-overall}
    \resizebox{\textwidth}{!}{\begin{tabular}{lcccccccccccc}\toprule
        & &\multicolumn{11}{c}{\#Layers in $\text{GNN}_B$ (256 hidden dims, aggregators: [max, mean, sum])} \\
        & &\multicolumn{5}{c}{\texttt{ogbg-molpcba}} & &\multicolumn{5}{c}{\texttt{ogbg-code2}} \\\cmidrule{3-7}\cmidrule{9-13}
        & &2 &4 &6 &8 &10 & &2 &4 &6 &8 &10 \\\hline
        \multirow{5}{*}{\makecell[l]{\#Layers in $\text{GNN}_A$\\ (256 hidden dims,\\~\![max, mean, sum])}} &2 &\cellcolor[HTML]{fdfdfd}-0.04 &\cellcolor[HTML]{f6c6de}\phantom{-}1.29 &\cellcolor[HTML]{f5bfda}\phantom{-}1.46 &\cellcolor[HTML]{f2accf}\phantom{-}1.88 &\cellcolor[HTML]{f1a8cc}\phantom{-}1.97 & &\cellcolor[HTML]{fefefe}-0.01 &\cellcolor[HTML]{f9d6e7}\phantom{-}0.31 &\cellcolor[HTML]{f8cfe3}\phantom{-}0.36 &\cellcolor[HTML]{f2acce}\phantom{-}0.63 &\cellcolor[HTML]{f1a8cc}\phantom{-}0.65 \\
        &4 &\cellcolor[HTML]{c0dbc9}-1.31 &\cellcolor[HTML]{fffeff}\phantom{-}0.03 &\cellcolor[HTML]{fadceb}\phantom{-}0.80 &\cellcolor[HTML]{f6c2db}\phantom{-}1.40 &\cellcolor[HTML]{f3b2d2}\phantom{-}1.75 & &\cellcolor[HTML]{dbeae0}-0.25 &\phantom{-}0.01 &\cellcolor[HTML]{f8d3e5}\phantom{-}0.34 &\cellcolor[HTML]{f7cae0}\phantom{-}0.40 &\cellcolor[HTML]{f4bad7}\phantom{-}0.52 \\
        &6 &\cellcolor[HTML]{b9d7c2}-1.48 &\cellcolor[HTML]{d7e8dc}-0.84 &\cellcolor[HTML]{fefefe}-0.01 &\cellcolor[HTML]{fbe2ee}\phantom{-}0.67 &\cellcolor[HTML]{fadceb}\phantom{-}0.81 & &\cellcolor[HTML]{cce2d3}-0.36 &\cellcolor[HTML]{d3e6d9}-0.31 &\cellcolor[HTML]{fefefe}-0.00 &\cellcolor[HTML]{fae0ed}\phantom{-}0.24 &\cellcolor[HTML]{f8d2e5}\phantom{-}0.34 \\
        &8 &\cellcolor[HTML]{afd1ba}-1.69 &\cellcolor[HTML]{c2dcca}-1.28 &\cellcolor[HTML]{ebf4ee}-0.41 &\cellcolor[HTML]{fffeff}\phantom{-}0.03 &\cellcolor[HTML]{fce7f1}\phantom{-}0.56 & &\cellcolor[HTML]{aacfb6}-0.59 &\cellcolor[HTML]{c6dfce}-0.40 &\cellcolor[HTML]{d9e9de}-0.27 &\phantom{-}0.00 &\cellcolor[HTML]{fce7f1}\phantom{-}0.19 \\
        &10 &\cellcolor[HTML]{a0c9ad}-2.01 &\cellcolor[HTML]{accfb7}-1.75 &\cellcolor[HTML]{cce2d3}-1.08 &\cellcolor[HTML]{dfede3}-0.67 &\cellcolor[HTML]{fefefe}-0.00 & &\cellcolor[HTML]{a0c9ad}-0.67 &\cellcolor[HTML]{bdd9c6}-0.46 &\cellcolor[HTML]{cde3d4}-0.35 &\cellcolor[HTML]{e5f0e8}-0.18 &\cellcolor[HTML]{f4f8f5}-0.08 \\
        \toprule
        & &\multicolumn{11}{c}{Hidden dims in $\text{GNN}_B$ (4 layers, aggregators: [max, mean, sum])} \\
        & &\multicolumn{5}{c}{\texttt{ogbg-molpcba}} & &\multicolumn{5}{c}{\texttt{ogbg-code2}} \\\cmidrule{3-7}\cmidrule{9-13}
        & &16 &32 &64 &128 &256 & &16 &32 &64 &128 &256 \\\hline
        \multirow{5}{*}{\makecell[l]{Hidden dims in $\text{GNN}_A$\\ (4 layers, [max, mean, sum])}} &16 &\phantom{-}0.02 &\cellcolor[HTML]{f8d0e4}\phantom{-}1.39 &\cellcolor[HTML]{f5bfda}\phantom{-}1.88 &\cellcolor[HTML]{f3afd0}\phantom{-}2.36 &\cellcolor[HTML]{f1a8cc}\phantom{-}2.55 & &\phantom{-}0.01 &\cellcolor[HTML]{f6c6de}\phantom{-}0.61 &\cellcolor[HTML]{f3b4d3}\phantom{-}0.81 &\cellcolor[HTML]{f3afd0}\phantom{-}0.86 &\cellcolor[HTML]{f1a8cc}\phantom{-}0.93 \\
        &32 &\cellcolor[HTML]{d0e4d6}-1.24 &\phantom{-}0.00 &\cellcolor[HTML]{fae0ed}\phantom{-}0.93 &\cellcolor[HTML]{f7c9df}\phantom{-}1.61 &\cellcolor[HTML]{f4b8d6}\phantom{-}2.09 & &\cellcolor[HTML]{bad8c3}-0.66 &\cellcolor[HTML]{fbfcfb}-0.04 &\cellcolor[HTML]{f7cee2}\phantom{-}0.53 &\cellcolor[HTML]{f4bad7}\phantom{-}0.74 &\cellcolor[HTML]{f2adcf}\phantom{-}0.88 \\
        &64 &\cellcolor[HTML]{a9ceb5}-2.29 &\cellcolor[HTML]{ddebe2}-0.89 &\phantom{-}0.02 &\cellcolor[HTML]{f9d6e7}\phantom{-}1.21 &\cellcolor[HTML]{f6c2db}\phantom{-}1.81 & &\cellcolor[HTML]{b3d3bd}-0.73 &\cellcolor[HTML]{bad8c3}-0.66 &\cellcolor[HTML]{fefefe}-0.00 &\cellcolor[HTML]{fadbea}\phantom{-}0.39 &\cellcolor[HTML]{f6c5dd}\phantom{-}0.62 \\
        &128 &\cellcolor[HTML]{a1caae}-2.49 &\cellcolor[HTML]{c2dccb}-1.61 &\cellcolor[HTML]{d9e9de}-1.00 &\cellcolor[HTML]{fefefe}-0.01 &\cellcolor[HTML]{f7cde2}\phantom{-}1.49 & &\cellcolor[HTML]{aacfb6}-0.81 &\cellcolor[HTML]{b4d4be}-0.72 &\cellcolor[HTML]{d4e6da}-0.41 &\cellcolor[HTML]{fcfdfc}-0.03 &\cellcolor[HTML]{f9d8e9}\phantom{-}0.42 \\
        &256 &\cellcolor[HTML]{a0c9ad}-2.54 &\cellcolor[HTML]{b2d3bd}-2.04 &\cellcolor[HTML]{bfdac8}-1.70 &\cellcolor[HTML]{cde2d4}-1.33 &\cellcolor[HTML]{fefefe}-0.01 & &\cellcolor[HTML]{a0c9ad}-0.91 &\cellcolor[HTML]{add0b8}-0.78 &\cellcolor[HTML]{bcd9c5}-0.64 &\cellcolor[HTML]{d5e7db}-0.40 &\cellcolor[HTML]{fcfdfc}-0.03 \\
        \toprule
    \end{tabular}}
    \resizebox{\textwidth}{!}{\begin{tabular}{lcx{1.05cm}x{1.05cm}x{1.05cm}x{1.05cm}cx{1.05cm}x{1.05cm}x{1.05cm}x{1.05cm}}
        & &\multicolumn{9}{c}{Aggregators in $\text{GNN}_B$ (4 layers, 64 hidden dims)} \\
        & &\multicolumn{4}{c}{\texttt{ogbg-molpcba}} & &\multicolumn{4}{c}{\texttt{ogbg-code2}} \\\cmidrule{3-6}\cmidrule{8-11}
        & &[max] &[mean] &[sum] &Comb. & &[max] &[mean] &[sum] &Comb. \\\hline
        \multirow{4}{*}{\makecell[l]{Aggregators in $\text{GNN}_A$\\ (4 layers, 64 hidden dims)}} &[max] &\cellcolor[HTML]{f7faf8}-0.03 &\cellcolor[HTML]{f8d1e4}\phantom{-}0.18 &\cellcolor[HTML]{f3b2d2}\phantom{-}0.31 &\cellcolor[HTML]{f1a8cc}0.34 & &\cellcolor[HTML]{fbfcfb}-0.02 &\cellcolor[HTML]{f9daea}\phantom{-}0.14 &\cellcolor[HTML]{f3b2d2}\phantom{-}0.30 &\cellcolor[HTML]{f1a8cc}\phantom{-}0.34 \\
        &[mean] &\cellcolor[HTML]{cce2d3}-0.17 &\cellcolor[HTML]{fcfdfd}-0.01 &\cellcolor[HTML]{f6c2dc}\phantom{-}0.24 &\cellcolor[HTML]{f2accf}0.33 & &\cellcolor[HTML]{ddebe1}-0.15 &\cellcolor[HTML]{fefefe}-0.00 &\cellcolor[HTML]{f8d3e5}\phantom{-}0.17 &\cellcolor[HTML]{f7cbe1}\phantom{-}0.20 \\
        &[sum] &\cellcolor[HTML]{a7cdb3}-0.30 &\cellcolor[HTML]{b7d6c1}-0.25 &\cellcolor[HTML]{fef4f8}\phantom{-}0.05 &\cellcolor[HTML]{f6c6de}0.23 & &\cellcolor[HTML]{bad7c3}-0.30 &\cellcolor[HTML]{e2eee6}-0.12 &\cellcolor[HTML]{fcfdfd}-0.01 &\cellcolor[HTML]{f9d8e8}\phantom{-}0.15 \\
        &Comb. &\cellcolor[HTML]{a0c9ad}-0.33 &\cellcolor[HTML]{a9ceb5}-0.29 &\cellcolor[HTML]{b9d7c3}-0.24 &\cellcolor[HTML]{fffbfd}0.02 & &\cellcolor[HTML]{a0c9ad}-0.41 &\cellcolor[HTML]{bdd9c6}-0.28 &\cellcolor[HTML]{dbebe0}-0.15 &\cellcolor[HTML]{fefefe}-0.00 \\
        \toprule
    \end{tabular}}
    \resizebox{\textwidth}{!}{\begin{tabular}{lcx{1cm}x{1cm}x{1cm}cccccccc}
        \multirow{3}{*}{\makecell{\texttt{ogbg-}\\\texttt{molpcba}}}& &\multicolumn{3}{c}{Edge feat. $-$ No edge feat.} & &\multicolumn{7}{c}{GNN architecture (4 layers, 64 hidden dims)} \\
        & &\multicolumn{3}{c}{Hidden dims (PNA)} & &\multicolumn{3}{c}{\texttt{ogbg-molpcba}} & &\multicolumn{3}{c}{\texttt{ogbg-code2}} \\\cmidrule{3-5}\cmidrule{7-9}\cmidrule{11-13}
        & &64 &128 &256 & &GCN &GIN &PNA & &GCN &GIN &PNA \\\hline
        \multirow{3}{*}{\makecell[l]{\#Layers\\ (PNA)}} &4 &\cellcolor[HTML]{a7cdb3}-0.42 &\cellcolor[HTML]{b7d6c1}-0.34 &\cellcolor[HTML]{d5e7db}-0.20 &GCN &\cellcolor[HTML]{f3f8f5}-0.07 &\cellcolor[HTML]{f5c1db}\phantom{-}0.34 &\cellcolor[HTML]{f1a8cc}\phantom{-}0.47 & &\cellcolor[HTML]{fefefe}-0.02 &\cellcolor[HTML]{f7c8df}\phantom{-}1.06 &\cellcolor[HTML]{f1a8cc}\phantom{-}1.66 \\
        &6 &\cellcolor[HTML]{a0c9ad}-0.45 &\cellcolor[HTML]{c3ddcb}-0.28 &\cellcolor[HTML]{d2e5d8}-0.21 &GIN &\cellcolor[HTML]{c9e0d0}-0.34 &\cellcolor[HTML]{fafcfb}-0.03 &\cellcolor[HTML]{f2adcf}\phantom{-}0.44 & &\cellcolor[HTML]{b7d6c1}-1.19 &\cellcolor[HTML]{fefefe}-0.01 &\cellcolor[HTML]{f4b6d4}\phantom{-}1.40 \\
        &8 &\cellcolor[HTML]{a5ccb2}-0.42 &\cellcolor[HTML]{c8dfcf}-0.26 &\cellcolor[HTML]{d4e6d9}-0.20 &PNA &\cellcolor[HTML]{a0c9ad}-0.59 &\cellcolor[HTML]{bbd8c5}-0.42 &\cellcolor[HTML]{fffcfe}\phantom{-}0.02 & &\cellcolor[HTML]{a0c9ad}-1.58 &\cellcolor[HTML]{a9ceb5}-1.43 &\cellcolor[HTML]{fefefe}-0.00 \\[-\aboverulesep]
        \bottomrule
    \end{tabular}}
\end{table*}

Training took from 1.7 hours (2-layer PNA vs. 2-layer PNA) to 4.2 hours (10-layer PNA vs. 10-layer PNA). The details of the training process are described in Appendix~\ref{sec:training-outcomes}. The training outcomes indicate that the more expressive GNN can continuously achieve a lower loss in our framework, and that GraphAC can successfully avoid information collapse.

The results of the experiments, recorded as $\mathcal{L}_{\text{GNN (edge features)}}-\mathcal{L}_{\text{GNN (no edge features)}}$ for the edge features group and $\mathcal{L}_{\text{GNN}_A}-\mathcal{L}_{\text{GNN}_B}$ for all other groups, are reported in Table~\ref{tab:results-overall}. Another two tables containing detailed results of the experiments can be found in Appendix~\ref{sec:results-details}. These results demonstrate that \emph{GraphAC can successfully distinguish GNNs of different expressiveness across various aspects, and consistently favors the more expressive GNNs}: 1)~deeper GNNs; 2)~GNNs with larger hidden dimensions; 3)~$\text{combining multiple aggregators}>\text{sum}>\text{mean}>\text{max}$ as aggregators \citep{Xu2019GIN,Corso2020PNA}; 4)~$\text{PNA}>\text{GIN}>\text{GCN}$ \citep{Xu2019GIN,Corso2020PNA}; and 5)~GNNs that include edge features. Furthermore, regardless of the ordering of the GNNs in the framework, there is no distinction between $\text{GNN}_A$ or $\text{GNN}_B$; this means that for all pairs of experiments, even if the order is switched, the absolute loss differences still remain roughly equal but only with the sign flipped. These observations suggest that \emph{GraphAC can genuinely distinguish different GNNs, regardless of their ordering in the framework}. Additionally, for the experiments with the same GNNs competing, GraphAC produced loss differences close to zero, demonstrating its ability to allow GNNs with the same expressiveness to tie, rather than falsely deciding a winner. Moreover, for every three GNNs: $\text{GNN}_A$, $\text{GNN}_B$ and $\text{GNN}_C$ in the experiments, if their expressiveness can be ordered as $\text{GNN}_A>\text{GNN}_B>\text{GNN}_C$, then there is also $|\mathcal{L}_{\text{GNN}_A}-\mathcal{L}_{\text{GNN}_C}|>|\mathcal{L}_{\text{GNN}_A}-\mathcal{L}_{\text{GNN}_B}|$ produced by GraphAC. This phenomenon shows that \emph{GraphAC is able to produce a total ordering of all GNNs}, which further demonstrates its credibility in GNN evaluation.

Some notable group-specific observations are as follows: 

\textbf{Different aggregators\ }\ It is observed that that the loss differences in the aggregators group are less significant than in the numbers of GNN layers and hidden dimensions groups. This is possibly because the effect on expressiveness by the aggregators is less significant than the number of parameters (i.e., number of layers and hidden dimensions).

\textbf{Different GNN architectures\ }\ It is also observed that the differences between architectures have a greater impact than simply changing the aggregators. This can be due to other differences, such as message passing framework in PNA compared to convolutions in GCN and GIN, and the added $\epsilon$ term in GIN compared to GCN.

\textbf{Inclusion of edge features\ }\ It is also observed that when the number of layers and hidden dimensions are larger, the magnitude of the loss difference becomes smaller. This is possibly because when a GNN is more complex, it can capture enough information from graphs even when they contain no edge information, thus the gain in performance by including edge features becomes relatively smaller.

\begin{figure}[t]
    \centering
    \begin{subfigure}[t]{0.4\textwidth}
        \includegraphics[scale=0.4, valign=t]{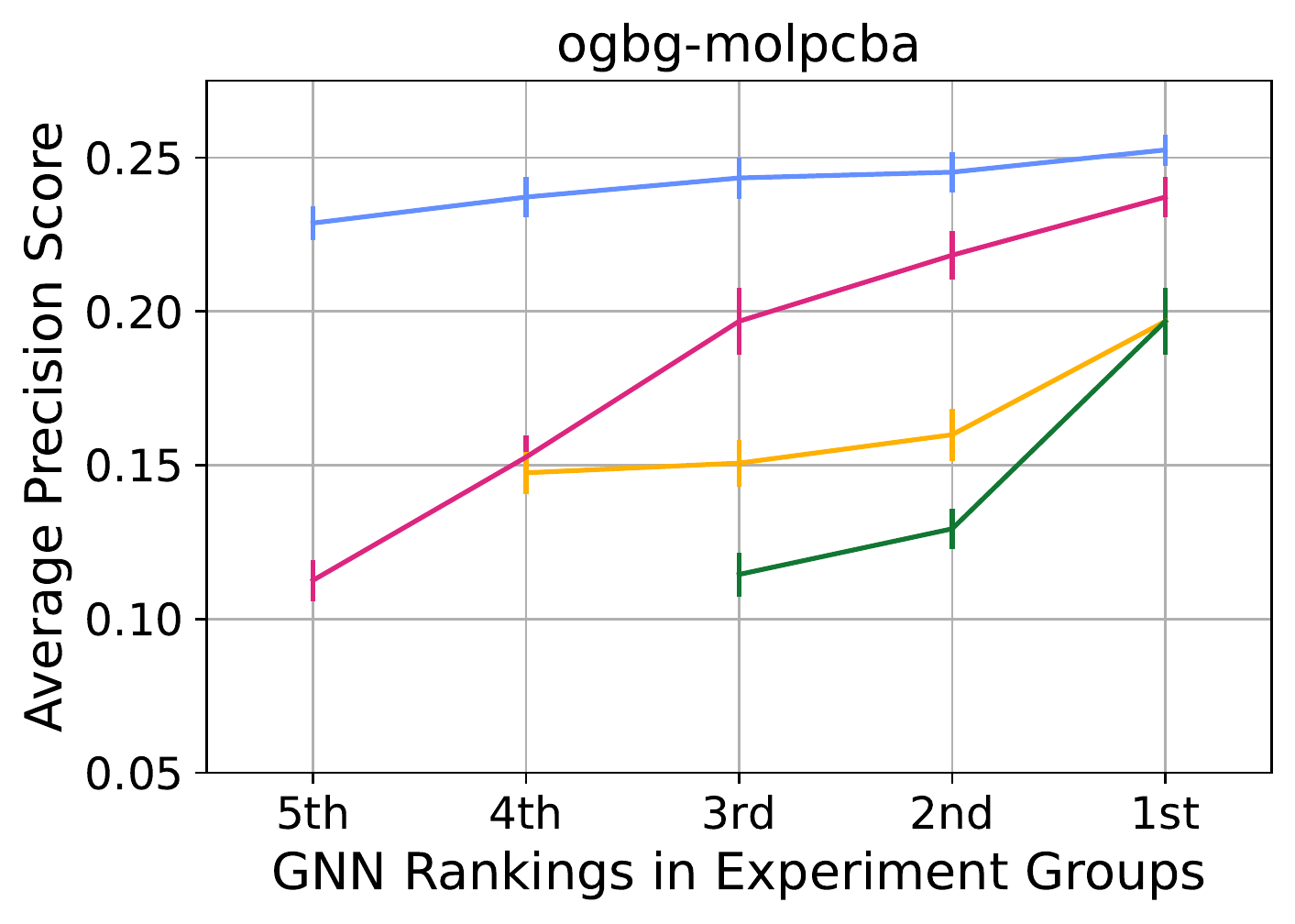}
    \end{subfigure}
    \begin{subfigure}[t]{0.59\textwidth}
        \includegraphics[scale=0.4, valign=t]{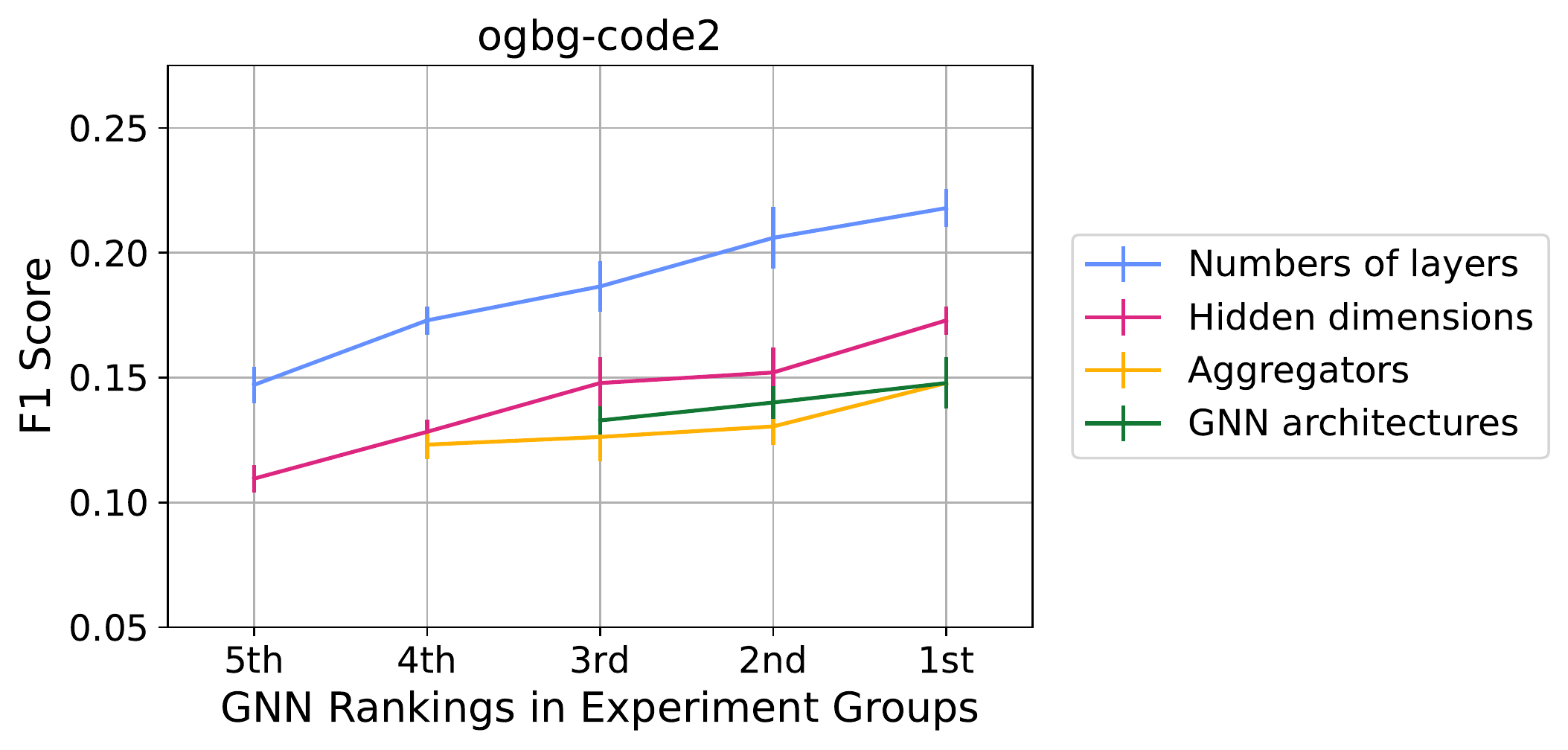}
    \end{subfigure}
    \caption{Correlation plots of GraphAC's GNN rankings with the GNNs' performances. The positive correlations show that GraphAC's expressiveness rankings align well with the GNNs' performances.}
    \label{fig:correlation}
    \vspace*{-\baselineskip}
\end{figure}

\subsection{Correlation with Task Performance}

In order to fully validate that the GNNs favored by GraphAC are indeed more expressive, we further evaluate all the GNNs used in the aforementioned experiments against the supervised learning tasks under the \texttt{ogbg-molpcba} and \texttt{ogbg-code2} datasets \citep{Hu2020OGB}. Each of the supervised training experiments takes 50 epochs. We then perform a correlation study on the GNNs' task performances with their expressiveness rankings produced by GraphAC.

Figure~\ref{fig:correlation} shows the GNNs' task performance on both datasets with respect to their expressiveness rankings produced by GraphAC, measured using the average precision of multi-class classification for the tasks on the \texttt{ogbg-molpcba} dataset, and F1 score of sub-token prediction for the tasks on the \texttt{ogbg-molpcba} dataset. The plots are divided into the experiment groups as presented in the previous section, in order to accurately demonstrate the correlation. The consistent, monotonic upward trend shows a strong correlation between the GNNs' task performance and GraphAC's expressiveness rankings on them, suggesting that \emph{GraphAC can genuinely distinguish GNNs of different expressiveness across various aspects, favoring the more expressive GNNs}. Separated plots of Figure~\ref{fig:correlation}, with one diagram per experiment group and detailed descriptions of the GNN architectures and parameters, can be found in Appendix~\ref{sec:correlation-plots}.

%% file: sections/conclusion.tex
\section{Conclusion}

We propose GraphAC (Graph Adversarial Collaboration), a novel, principled, and task-agnostic framework for evaluating GNNs through contrastive self-supervision, without the need of handcrafted augmentations. Inspired by the Barlow Twins loss \citep{Zbontar2021BarlowTwins}, we introduce a novel objective function: the Competitive Barlow Twins, which replaces its redundancy reduction term with a difference between the upper-triangle and lower-triangle of the cross-correlation matrix of the two GNN's output embeddings. GraphAC successfully distinguishes GNNs of different expressiveness in all experiments within graphs of two distinct contexts: molecular graphs and abstract syntax trees, across various aspects including the number of layers, hidden dimensionality, aggregators, GNN architecture and edge features, and ensures that more expressive GNNs can always win with a statistically significant difference. GraphAC is also able to estimate the degree of expressiveness of different GNNs, and produce a total ordering of all GNNs with its measurements. GraphAC provides a novel principle of evaluating GNNs and an effective contrastive SSL framework without requiring any augmentations, making a notable contribution to the graph SSL and molecular representation learning community, which can be applied to many important tasks in drug discovery.

We believe that the success of GraphAC opens up a new, principled way of thinking when developing contrastive SSL methods, by considering the more expressive GNN as an encoder that learns more complex but less general information from the graphs, and the less expressive GNN as one that captures more basic but general information. Consequently, combining the two GNNs creates a better overall understanding of the graphs and can be used to perform SSL on graphs without manually applying augmentations, which may have introduced arbitrary human knowledge that were not originally provided by the training data.

%% file: sections/acknowledgements.tex
\section*{Acknowledgements}

This work was performed using resources provided by the Cambridge Service for Data Driven Discovery (CSD3) operated by the University of Cambridge Research Computing Service, provided by Dell EMC and Intel using Tier-2 funding from the Engineering and Physical Sciences Research Council (capital grant EP/T022159/1), and DiRAC funding from the Science and Technology Facilities Council. For the purpose of open access, the authors have applied a Creative Commons Attribution (CC BY) licence to any Author Accepted Manuscript version arising.

%% file: appendices/algorithm.tex
\section{Algorithm} \label{sec:pseudocode}

\begin{algorithm}[!htbp]
\caption{PyTorch-style pseudocode for GraphAC's framework}
\begin{lstlisting}[language=myPython]
# gnn_a, gnn_b: GNN encoder networks
# alpha, beta, lambd, mu: coefficients of the loss terms
# N: batch size
# d: dimensionality of the embeddings
#
# diagonal: on-diagonal elements of a matrix
# off_diagonal: off_diagonal elements of a matrix
# triu: upper-triangle elements of a matrix
# tril: lower-triangle elements of a matrix

for x in dataloader: # load a batch with N samples
    # compute embeddings
    h_a_out = gnn_a(x)   # N x d
    h_b_out = gnn_b(x)   # N x d
    
    # normalize embeddings along the batch dimension (VICReg)
    h_a_out_norm = (h_a_out - h_a_out.mean(dim=0))  # N x d
    h_b_out_norm = (h_b_out - h_b_out.mean(dim=0))  # N x d
    
    # covariance matrices
    cov_a = (h_a_out_norm.T @ h_a_out_norm) / (N - 1)  # d x d
    cov_b = (h_b_out_norm.T @ h_b_out_norm) / (N - 1)  # d x d
    
    # covariance regularisation loss
    cov_loss = off_diagonal(cov_a).pow(2).sum() / d \
               + off_diagonal(cov_b).pow(2).sum() / d
    
    # normalize embeddings along the batch dimension (Barlow Twins)
    h_a_out_norm = h_a_out_norm / h_a_out.std(dim=0)  # N x d
    h_b_out_norm = h_a_out_norm / h_a_out.std(dim=0)  # N x d
    
    # cross-correlation matrix
    corr = (h_a_out_norm.T @ h_b_out_norm) / N  # d x d
    
    # Competitive Barlow Twins loss components
    on_diag = (diagonal(corr) - 1).pow(2).sum()
    upper_tri = triu(corr, diagonal=1).pow(2).sum()
    upper_tri = tril(corr, diagonal=-1).pow(2).sum()
              
    # Competitive Barlow Twins losses
    loss_a = alpha * (on_diag + lambd * (upper_tri - mu * lower_tri)) \
             + beta * cov_loss
    loss_b = alpha * (on_diag + lambd * (lower_tri - mu * upper_tri)) \
             + beta * cov_loss
    
    # optimisation steps
    loss_a.backward()
    loss_b.backward()
    optimiser_a.step()
    optimiser_b.step()
    optimiser_a.zero_grad()
    optimiser_b.zero_grad()
\end{lstlisting}
\end{algorithm}

\newpage

%% file: appendices/results-details.tex
\section{Detailed Results} \label{sec:results-details}

\begin{table*}[h]\centering
    \caption{Loss differences of the conducted experiments on the \texttt{ogbg-molpcba} dataset, with the uncertainty included and with a higher precision.}
    \label{tab:results-pcba}
    \resizebox{\textwidth}{!}{\begin{tabular}{lcccccc}\toprule
        & &\multicolumn{5}{c}{\#Layers in $\text{GNN}_B$ (256 hidden dims, aggregators: [max, mean, sum])} \\\cmidrule{3-7}
        & &2 &4 &6 &8 &10 \\\hline
        \multirow{5}{*}{\makecell[l]{\#Layers in $\text{GNN}_A$\\ (256 hidden dims,\\~\![max, mean, sum])}} &2 &\cellcolor[HTML]{fdfdfd}-0.042 $\pm$ 0.095 &\cellcolor[HTML]{f6c6de}\phantom{-}1.290 $\pm$ 0.127 &\cellcolor[HTML]{f5bfda}\phantom{-}1.458 $\pm$ 0.101 &\cellcolor[HTML]{f2accf}\phantom{-}1.882 $\pm$ 0.154 &\cellcolor[HTML]{f1a8cc}\phantom{-}1.966 $\pm$ 0.175 \\
        &4 &\cellcolor[HTML]{c0dbc9}-1.315 $\pm$ 0.106 &\cellcolor[HTML]{fffeff}\phantom{-}0.027 $\pm$ 0.097 &\cellcolor[HTML]{fadceb}\phantom{-}0.799 $\pm$ 0.190 &\cellcolor[HTML]{f6c2db}\phantom{-}1.399 $\pm$ 0.264 &\cellcolor[HTML]{f3b2d2}\phantom{-}1.748 $\pm$ 0.344 \\
        &6 &\cellcolor[HTML]{b9d7c2}-1.478 $\pm$ 0.218 &\cellcolor[HTML]{d7e8dc}-0.845 $\pm$ 0.320 &\cellcolor[HTML]{fefefe}-0.014 $\pm$ 0.069 &\cellcolor[HTML]{fbe2ee}\phantom{-}0.674 $\pm$ 0.211 &\cellcolor[HTML]{fadceb}\phantom{-}0.808 $\pm$ 0.152 \\
        &8 &\cellcolor[HTML]{afd1ba}-1.687 $\pm$ 0.284 &\cellcolor[HTML]{c2dcca}-1.276 $\pm$ 0.269 &\cellcolor[HTML]{ebf4ee}-0.406 $\pm$ 0.087 &\cellcolor[HTML]{fffeff}\phantom{-}0.030 $\pm$ 0.098 &\cellcolor[HTML]{fce7f1}\phantom{-}0.555 $\pm$ 0.176 \\
        &10 &\cellcolor[HTML]{a0c9ad}-2.008 $\pm$ 0.245 &\cellcolor[HTML]{accfb7}-1.751 $\pm$ 0.235 &\cellcolor[HTML]{cce2d3}-1.076 $\pm$ 0.298 &\cellcolor[HTML]{dfede3}-0.668 $\pm$ 0.160 &\cellcolor[HTML]{fefefe}-0.004 $\pm$ 0.065 \\
        \toprule
        & &\multicolumn{5}{c}{Hidden dims in $\text{GNN}_B$ (4 layers, aggregators: [max, mean, sum])} \\\cmidrule{3-7}
        & &16 &32 &64 &128 &256 \\\hline
        \multirow{5}{*}{\makecell[l]{Hidden dims in $\text{GNN}_A$\\ (4 layers, [max, mean, sum])}} &16 &\phantom{-}0.022 $\pm$ 0.078 &\cellcolor[HTML]{f8d0e4}\phantom{-}1.390 $\pm$ 0.301 &\cellcolor[HTML]{f5bfda}\phantom{-}1.883 $\pm$ 0.174 &\cellcolor[HTML]{f3afd0}\phantom{-}2.356 $\pm$ 0.238 &\cellcolor[HTML]{f1a8cc}\phantom{-}2.554 $\pm$ 0.293 \\
        &32 &\cellcolor[HTML]{d0e4d6}-1.240 $\pm$ 0.215 &\phantom{-}0.001 $\pm$ 0.083 &\cellcolor[HTML]{fae0ed}\phantom{-}0.932 $\pm$ 0.238 &\cellcolor[HTML]{f7c9df}\phantom{-}1.614 $\pm$ 0.287 &\cellcolor[HTML]{f4b8d6}\phantom{-}2.093 $\pm$ 0.301 \\
        &64 &\cellcolor[HTML]{a9ceb5}-2.290 $\pm$ 0.155 &\cellcolor[HTML]{ddebe2}-0.895 $\pm$ 0.228 &\phantom{-}0.018 $\pm$ 0.047 &\cellcolor[HTML]{f9d6e7}\phantom{-}1.206 $\pm$ 0.229 &\cellcolor[HTML]{f6c2db}\phantom{-}1.813 $\pm$ 0.246 \\
        &128 &\cellcolor[HTML]{a1caae}-2.493 $\pm$ 0.262 &\cellcolor[HTML]{c2dccb}-1.609 $\pm$ 0.340 &\cellcolor[HTML]{d9e9de}-0.997 $\pm$ 0.166 &\cellcolor[HTML]{fefefe}-0.010 $\pm$ 0.084 &\cellcolor[HTML]{f7cde2}\phantom{-}1.490 $\pm$ 0.143 \\
        &256 &\cellcolor[HTML]{a0c9ad}-2.541 $\pm$ 0.235 &\cellcolor[HTML]{b2d3bd}-2.041 $\pm$ 0.348 &\cellcolor[HTML]{bfdac8}-1.704 $\pm$ 0.234 &\cellcolor[HTML]{cde2d4}-1.330 $\pm$ 0.212 &\cellcolor[HTML]{fefefe}-0.006 $\pm$ 0.096 \\
        \toprule
    \end{tabular}}
    \resizebox{\textwidth}{!}{\begin{tabular}{lcx{2.5cm}x{2.5cm}x{2.5cm}x{2.5cm}}
        & &\multicolumn{4}{c}{Aggregators in $\text{GNN}_B$ (4 layers, 64 hidden dims)} \\\cmidrule{3-6}
        & &[max] &[mean] &[sum] &[max, mean, sum] \\\hline
        \multirow{4}{*}{\makecell[l]{Aggregators in $\text{GNN}_A$\\ (4 layers, 64 hidden dims)}} &[max] &\cellcolor[HTML]{f7faf8}-0.026 $\pm$ 0.060 &\cellcolor[HTML]{f8d1e4}\phantom{-}0.182 $\pm$ 0.091 &\cellcolor[HTML]{f3b2d2}\phantom{-}0.305 $\pm$ 0.059 &\cellcolor[HTML]{f1a8cc}0.342 $\pm$ 0.051 \\
        &[mean] &\cellcolor[HTML]{cce2d3}-0.174 $\pm$ 0.041 &\cellcolor[HTML]{fcfdfd}-0.007 $\pm$ 0.040 &\cellcolor[HTML]{f6c2dc}\phantom{-}0.241 $\pm$ 0.068 &\cellcolor[HTML]{f2accf}0.328 $\pm$ 0.069 \\
        &[sum] &\cellcolor[HTML]{a7cdb3}-0.304 $\pm$ 0.032 &\cellcolor[HTML]{b7d6c1}-0.248 $\pm$ 0.038 &\cellcolor[HTML]{fef4f8}\phantom{-}0.047 $\pm$ 0.028 &\cellcolor[HTML]{f6c6de}0.228 $\pm$ 0.049 \\
        &[max,mean,sum] &\cellcolor[HTML]{a0c9ad}-0.329 $\pm$ 0.068 &\cellcolor[HTML]{a9ceb5}-0.295 $\pm$ 0.047 &\cellcolor[HTML]{b9d7c3}-0.239 $\pm$ 0.045 &\cellcolor[HTML]{fffbfd}0.018 $\pm$ 0.047 \\
        \toprule
    \end{tabular}}
    \resizebox{\textwidth}{!}{\begin{tabular}{lccccccccc}
        & &\multicolumn{3}{c}{$\text{GNN}_B$ architecture (4 layers, 64 hidden dims)} & & &\multicolumn{3}{c}{Hidden dims (PNA, edge feat. $-$ no edge feat.)} \\\cmidrule{3-5}\cmidrule{8-10}
        & &GCN &GIN &PNA & & &64 &128 &256 \\\hline
        \multirow{3}{*}{$\text{GNN}_A$} &GCN &\cellcolor[HTML]{f3f8f5}-0.069 $\pm$ 0.073 &\cellcolor[HTML]{f5c1db}\phantom{-}0.338 $\pm$ 0.074 &\cellcolor[HTML]{f1a8cc}0.467 $\pm$ 0.117 &\multirow{3}{*}{\#Layers} &4 &\cellcolor[HTML]{a7cdb3}-0.418 $\pm$ 0.096 &\cellcolor[HTML]{b7d6c1}-0.340 $\pm$ 0.118 &\cellcolor[HTML]{d5e7db}-0.198 $\pm$ 0.060 \\
        &GIN &\cellcolor[HTML]{c9e0d0}-0.337 $\pm$ 0.056 &\cellcolor[HTML]{fafcfb}-0.026 $\pm$ 0.018 &\cellcolor[HTML]{f2adcf}0.441 $\pm$ 0.142 & &6 &\cellcolor[HTML]{a0c9ad}-0.453 $\pm$ 0.119 &\cellcolor[HTML]{c3ddcb}-0.285 $\pm$ 0.085 &\cellcolor[HTML]{d2e5d8}-0.214 $\pm$ 0.089 \\
        &PNA &\cellcolor[HTML]{a0c9ad}-0.594 $\pm$ 0.200 &\cellcolor[HTML]{bbd8c5}-0.419 $\pm$ 0.160 &\cellcolor[HTML]{fffcfe}0.018 $\pm$ 0.047 & &8 &\cellcolor[HTML]{a5ccb2}-0.425 $\pm$ 0.100 &\cellcolor[HTML]{c8dfcf}-0.261 $\pm$ 0.099 &\cellcolor[HTML]{d4e6d9}-0.205 $\pm$ 0.061 \\[-\aboverulesep]
        \bottomrule
    \end{tabular}}

    \bigskip
    \medskip

    \caption{Loss differences of the conducted experiments on the \texttt{ogbg-code2} dataset, with the uncertainty included and with a higher precision. Since this dataset does not contain any edge features, no experiments regarding the inclusion of edge features for this dataset are conducted.}
    \label{tab:results-code2}
    \resizebox{\textwidth}{!}{\begin{tabular}{lcccccc}\toprule
        & &\multicolumn{5}{c}{\#Layers in $\text{GNN}_B$ (256 hidden dims, aggregators: [max, mean, sum])} \\\cmidrule{3-7}
        & &2 &4 &6 &8 &10 \\\hline
        \multirow{5}{*}{\makecell[l]{\#Layers in $\text{GNN}_A$\\ (256 hidden dims,\\~\![max, mean, sum])}} &2 &\cellcolor[HTML]{fefefe}-0.006 $\pm$ 0.024 &\cellcolor[HTML]{f9d6e7}\phantom{-}0.312 $\pm$ 0.189 &\cellcolor[HTML]{f8cfe3}\phantom{-}0.363 $\pm$ 0.115 &\cellcolor[HTML]{f2acce}\phantom{-}0.629 $\pm$ 0.285 &\cellcolor[HTML]{f1a8cc}\phantom{-}0.655 $\pm$ 0.152 \\
        &4 &\cellcolor[HTML]{dbeae0}-0.249 $\pm$ 0.135 &\phantom{-}0.005 $\pm$ 0.027 &\cellcolor[HTML]{f8d3e5}\phantom{-}0.338 $\pm$ 0.091 &\cellcolor[HTML]{f7cae0}\phantom{-}0.404 $\pm$ 0.147 &\cellcolor[HTML]{f4bad7}\phantom{-}0.524 $\pm$ 0.115 \\
        &6 &\cellcolor[HTML]{cce2d3}-0.357 $\pm$ 0.177 &\cellcolor[HTML]{d3e6d9}-0.308 $\pm$ 0.137 &\cellcolor[HTML]{fefefe}-0.002 $\pm$ 0.018 &\cellcolor[HTML]{fae0ed}\phantom{-}0.239 $\pm$ 0.142 &\cellcolor[HTML]{f8d2e5}\phantom{-}0.342 $\pm$ 0.116 \\
        &8 &\cellcolor[HTML]{aacfb6}-0.594 $\pm$ 0.143 &\cellcolor[HTML]{c6dfce}-0.395 $\pm$ 0.111 &\cellcolor[HTML]{d9e9de}-0.267 $\pm$ 0.162 &\phantom{-}0.001 $\pm$ 0.024 &\cellcolor[HTML]{fce7f1}\phantom{-}0.185 $\pm$ 0.114 \\
        &10 &\cellcolor[HTML]{a0c9ad}-0.670 $\pm$ 0.385 &\cellcolor[HTML]{bdd9c6}-0.463 $\pm$ 0.133 &\cellcolor[HTML]{cde3d4}-0.346 $\pm$ 0.124 &\cellcolor[HTML]{e5f0e8}-0.183 $\pm$ 0.118 &\cellcolor[HTML]{f4f8f5}-0.077 $\pm$ 0.023 \\
        \toprule
        & &\multicolumn{5}{c}{Hidden dims in $\text{GNN}_B$ (4 layers, aggregators: [max, mean, sum])} \\\cmidrule{3-7}
        & &16 &32 &64 &128 &256 \\\hline
        \multirow{5}{*}{\makecell[l]{Hidden dims in $\text{GNN}_A$\\ (4 layers, [max, mean, sum])}} &16 &\phantom{-}0.008 $\pm$ 0.051 &\cellcolor[HTML]{f6c6de}\phantom{-}0.612 $\pm$ 0.134 &\cellcolor[HTML]{f3b4d3}\phantom{-}0.805 $\pm$ 0.185 &\cellcolor[HTML]{f3afd0}\phantom{-}0.856 $\pm$ 0.156 &\cellcolor[HTML]{f1a8cc}\phantom{-}0.926 $\pm$ 0.168 \\
        &32 &\cellcolor[HTML]{bad8c3}-0.660 $\pm$ 0.207 &\cellcolor[HTML]{fbfcfb}-0.036 $\pm$ 0.046 &\cellcolor[HTML]{f7cee2}\phantom{-}0.532 $\pm$ 0.195 &\cellcolor[HTML]{f4bad7}\phantom{-}0.742 $\pm$ 0.195 &\cellcolor[HTML]{f2adcf}\phantom{-}0.875 $\pm$ 0.290 \\
        &64 &\cellcolor[HTML]{b3d3bd}-0.728 $\pm$ 0.128 &\cellcolor[HTML]{bad8c3}-0.659 $\pm$ 0.233 &\cellcolor[HTML]{fefefe}-0.002 $\pm$ 0.043 &\cellcolor[HTML]{fadbea}\phantom{-}0.389 $\pm$ 0.085 &\cellcolor[HTML]{f6c5dd}\phantom{-}0.620 $\pm$ 0.155 \\
        &128 &\cellcolor[HTML]{aacfb6}-0.810 $\pm$ 0.166 &\cellcolor[HTML]{b4d4be}-0.716 $\pm$ 0.163 &\cellcolor[HTML]{d4e6da}-0.410 $\pm$ 0.115 &\cellcolor[HTML]{fcfdfc}-0.027 $\pm$ 0.056 &\cellcolor[HTML]{f9d8e9}\phantom{-}0.417 $\pm$ 0.156 \\
        &256 &\cellcolor[HTML]{a0c9ad}-0.914 $\pm$ 0.129 &\cellcolor[HTML]{add0b8}-0.784 $\pm$ 0.141 &\cellcolor[HTML]{bcd9c5}-0.636 $\pm$ 0.129 &\cellcolor[HTML]{d5e7db}-0.396 $\pm$ 0.147 &\cellcolor[HTML]{fcfdfc}-0.028 $\pm$ 0.068 \\
        \toprule
    \end{tabular}}
    \resizebox{\textwidth}{!}{\begin{tabular}{lcx{2.5cm}x{2.5cm}x{2.5cm}x{2.5cm}}
        & &\multicolumn{4}{c}{Aggregators in $\text{GNN}_B$ (4 layers, 64 hidden dims)} \\\cmidrule{3-6}
        & &[max] &[mean] &[sum] &[max, mean, sum] \\\hline
        \multirow{4}{*}{\makecell[l]{Aggregators in $\text{GNN}_A$\\ (4 layers, 64 hidden dims)}} &[max] &\cellcolor[HTML]{fbfcfb}-0.016 $\pm$ 0.052 &\cellcolor[HTML]{f9daea}\phantom{-}0.144 $\pm$ 0.067 &\cellcolor[HTML]{f3b2d2}\phantom{-}0.300 $\pm$ 0.087 &\cellcolor[HTML]{f1a8cc}\phantom{-}0.335 $\pm$ 0.148 \\
        &[mean] &\cellcolor[HTML]{ddebe1}-0.146 $\pm$ 0.071 &\cellcolor[HTML]{fefefe}-0.001 $\pm$ 0.045 &\cellcolor[HTML]{f8d3e5}\phantom{-}0.172 $\pm$ 0.089 &\cellcolor[HTML]{f7cbe1}\phantom{-}0.202 $\pm$ 0.074 \\
        &[sum] &\cellcolor[HTML]{bad7c3}-0.296 $\pm$ 0.199 &\cellcolor[HTML]{e2eee6}-0.124 $\pm$ 0.048 &\cellcolor[HTML]{fcfdfd}-0.009 $\pm$ 0.049 &\cellcolor[HTML]{f9d8e8}\phantom{-}0.152 $\pm$ 0.059 \\
        &[max,mean,sum] &\cellcolor[HTML]{a0c9ad}-0.409 $\pm$ 0.267 &\cellcolor[HTML]{bdd9c6}-0.283 $\pm$ 0.230 &\cellcolor[HTML]{dbebe0}-0.151 $\pm$ 0.110 &\cellcolor[HTML]{fefefe}-0.002 $\pm$ 0.043 \\
        \toprule
    \end{tabular}}
    \resizebox{\textwidth}{!}{\begin{tabular}{lx{2.1cm}x{3.33cm}x{3.33cm}x{3.33cm}}
        & &\multicolumn{3}{c}{$\text{GNN}_B$ architecture (4 layers, 64 hidden dims)} \\\cmidrule{3-5}
        & &GCN &GIN &PNA \\\hline
        \multirow{3}{*}{\makecell[l]{$\text{GNN}_A$ architecture\\ (4 layers, 64 hidden dims)}} &GCN &\cellcolor[HTML]{fefefe}-0.016 $\pm$ 0.056 &\cellcolor[HTML]{f7c8df}\phantom{-}1.061 $\pm$ 0.416 &\cellcolor[HTML]{f1a8cc}\phantom{-}1.662 $\pm$ 0.402 \\
        &GIN &\cellcolor[HTML]{b7d6c1}-1.192 $\pm$ 0.493 &\cellcolor[HTML]{fefefe}-0.013 $\pm$ 0.062 &\cellcolor[HTML]{f4b6d4}\phantom{-}1.403 $\pm$ 0.221 \\
        &PNA &\cellcolor[HTML]{a0c9ad}-1.582 $\pm$ 0.457 &\cellcolor[HTML]{a9ceb5}-1.427 $\pm$ 0.360 &\cellcolor[HTML]{fefefe}-0.002 $\pm$ 0.043 \\[-\aboverulesep]
        \bottomrule
    \end{tabular}}
\end{table*}

%% file: appendices/correlation-study.tex
\section{Detailed correlation plots} \label{sec:correlation-plots}

\subsection{ogbg-molpcba Dataset} \label{sec:correlation-plots-pcba}

\begin{figure}[h]
    \centering
    \hspace{2.5cm}
    \includegraphics[width=0.75\textwidth]{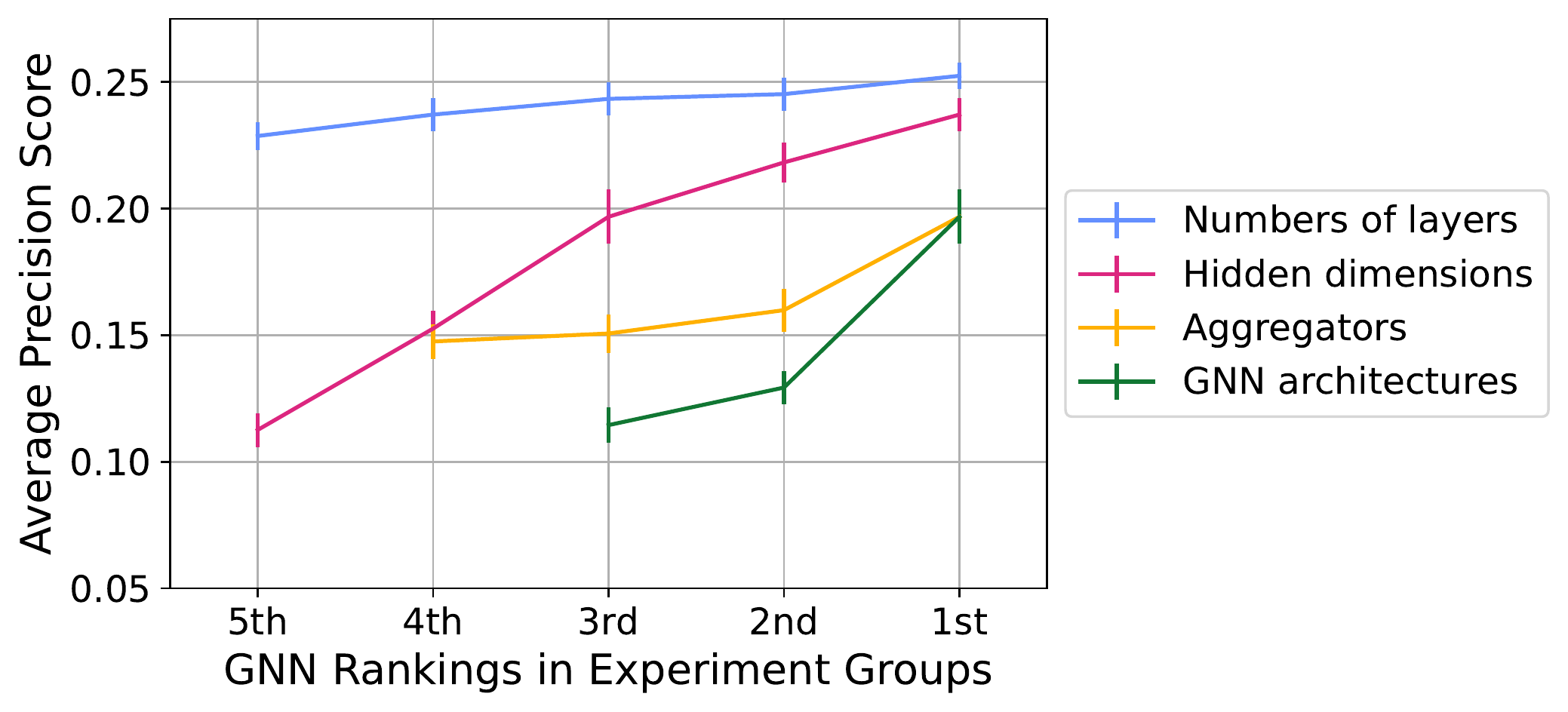}
\end{figure}

\begin{figure}[h]
    \centering
    \begin{subfigure}[t]{0.49\textwidth}
        \includegraphics[scale=0.45, valign=t]{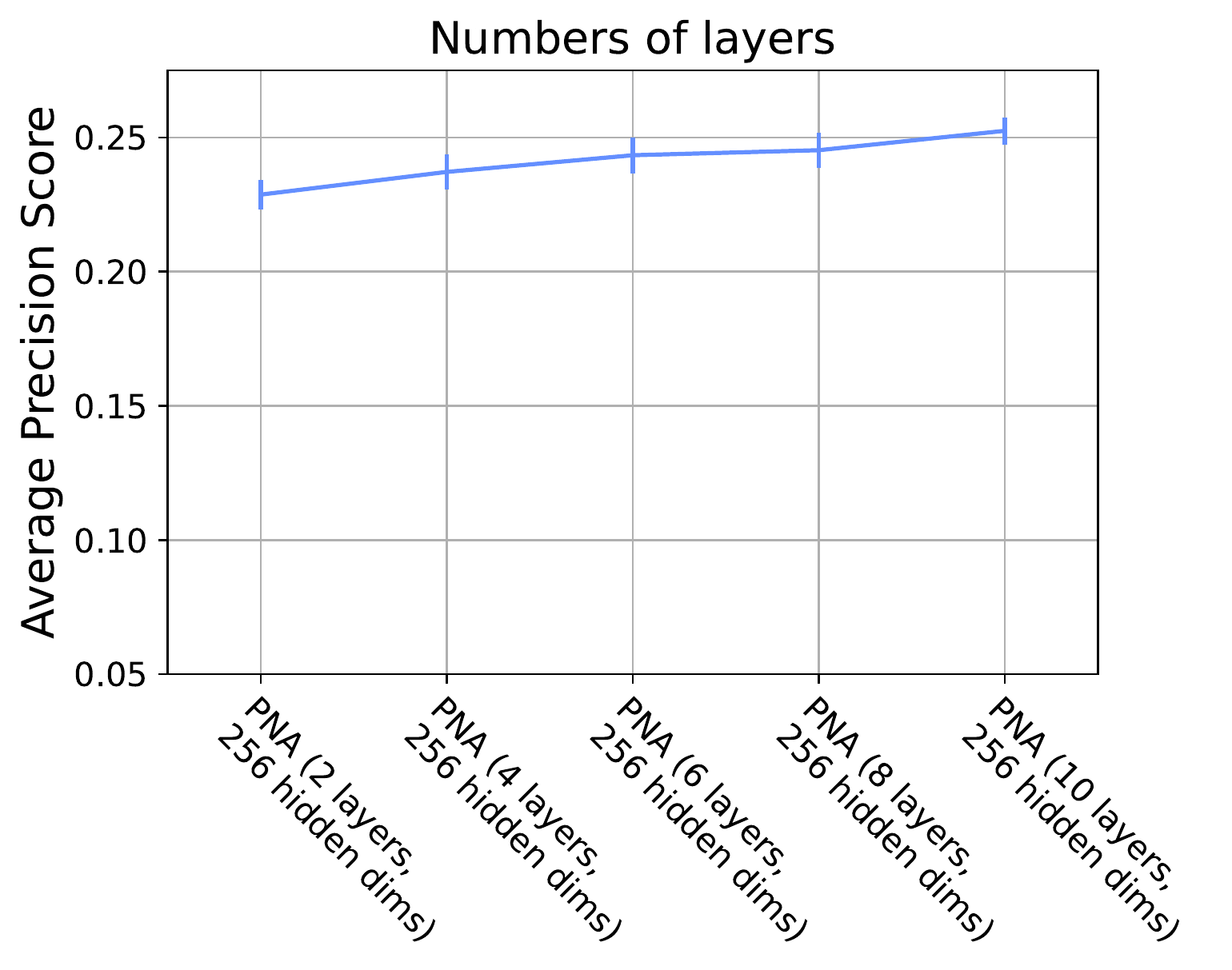}
    \end{subfigure}
    \begin{subfigure}[t]{0.49\textwidth}
        \includegraphics[scale=0.45, valign=t]{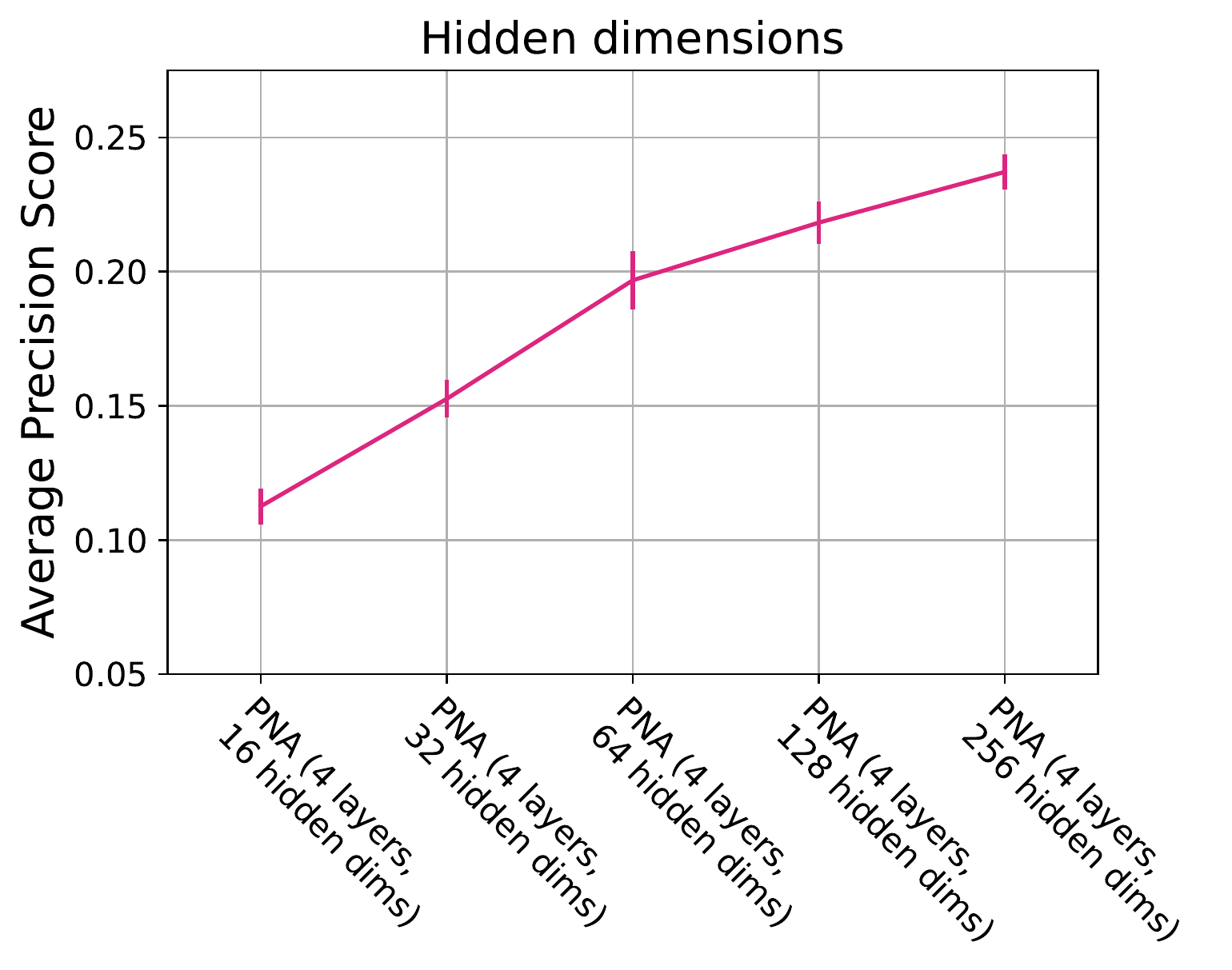}
    \end{subfigure}
    \begin{subfigure}[t]{0.49\textwidth}
        \includegraphics[scale=0.45, valign=t]{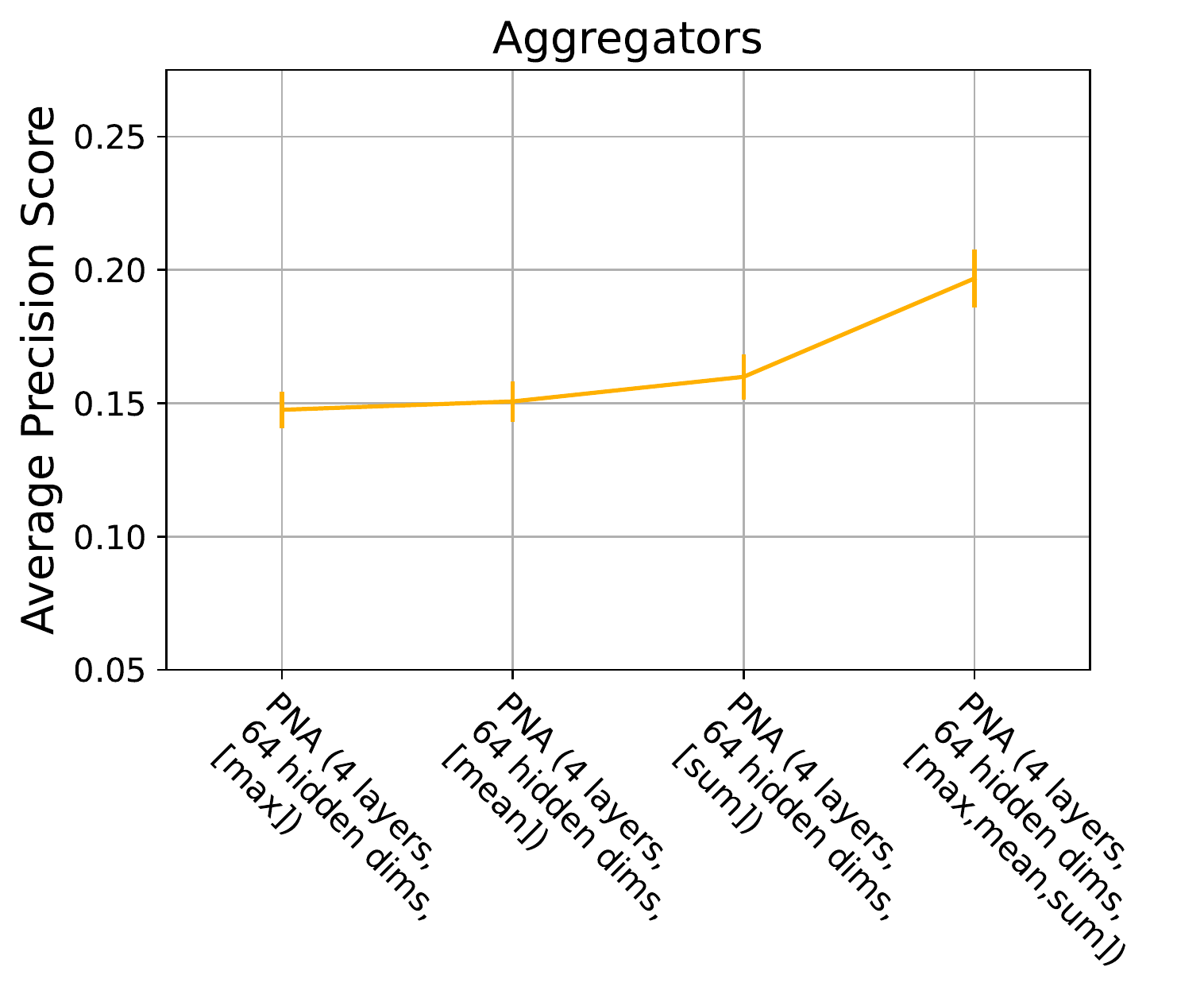}
    \end{subfigure}
    \begin{subfigure}[t]{0.49\textwidth}
        \includegraphics[scale=0.45, valign=t]{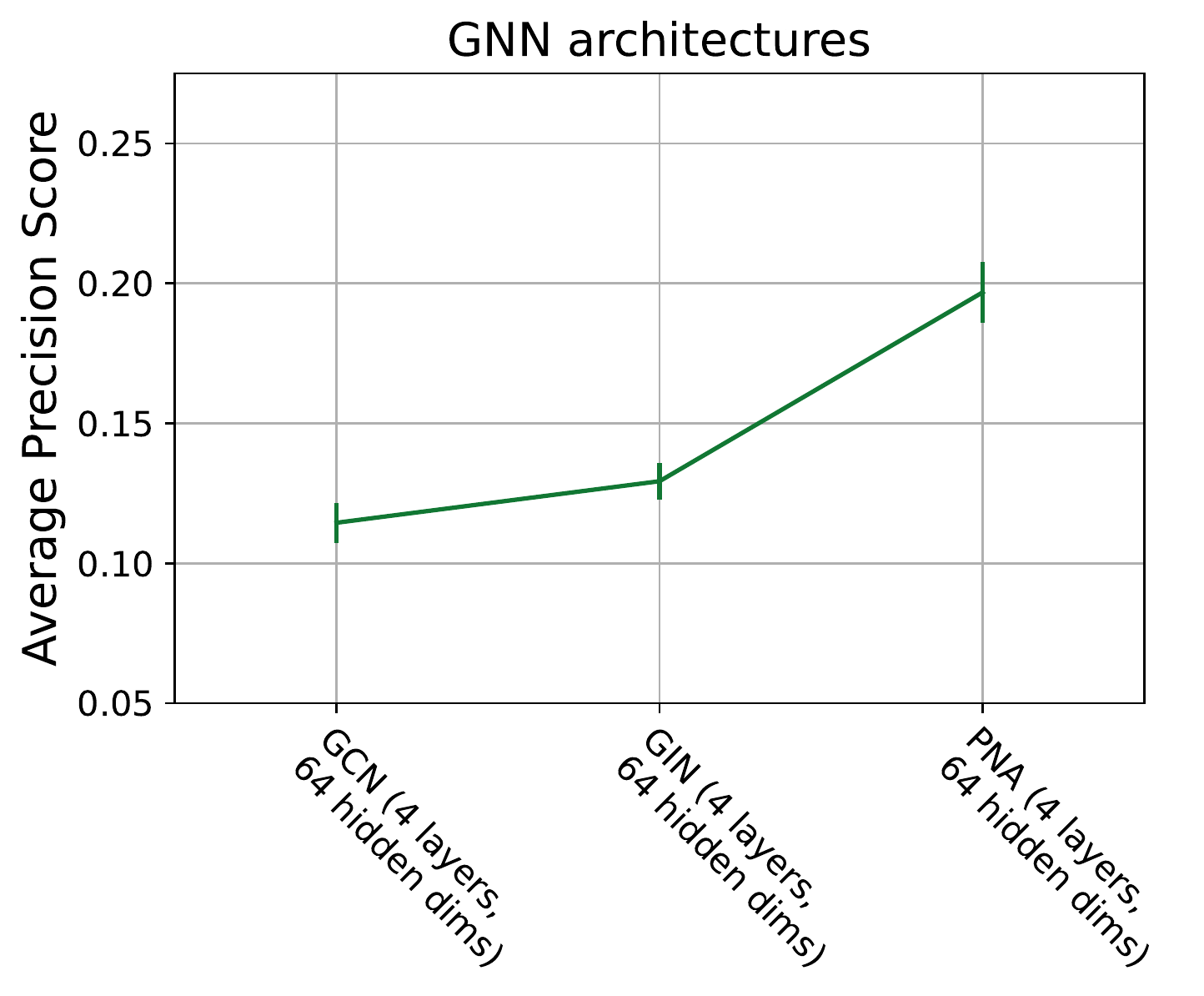}
    \end{subfigure}
    \caption{Correlation plots of GraphAC's GNN rankings with the GNNs' task performances on the \texttt{ogbg-molpcba} dataset, with detailed descriptions of the GNN architectures and parameters.}
    \label{fig:correlation-detailed-pcba}
\end{figure}

\newpage

\subsection{ogbg-code2 Dataset} \label{sec:correlation-plots-code2}

\begin{figure}[h]
    \centering
    \hspace{2.5cm}
    \includegraphics[width=0.75\textwidth]{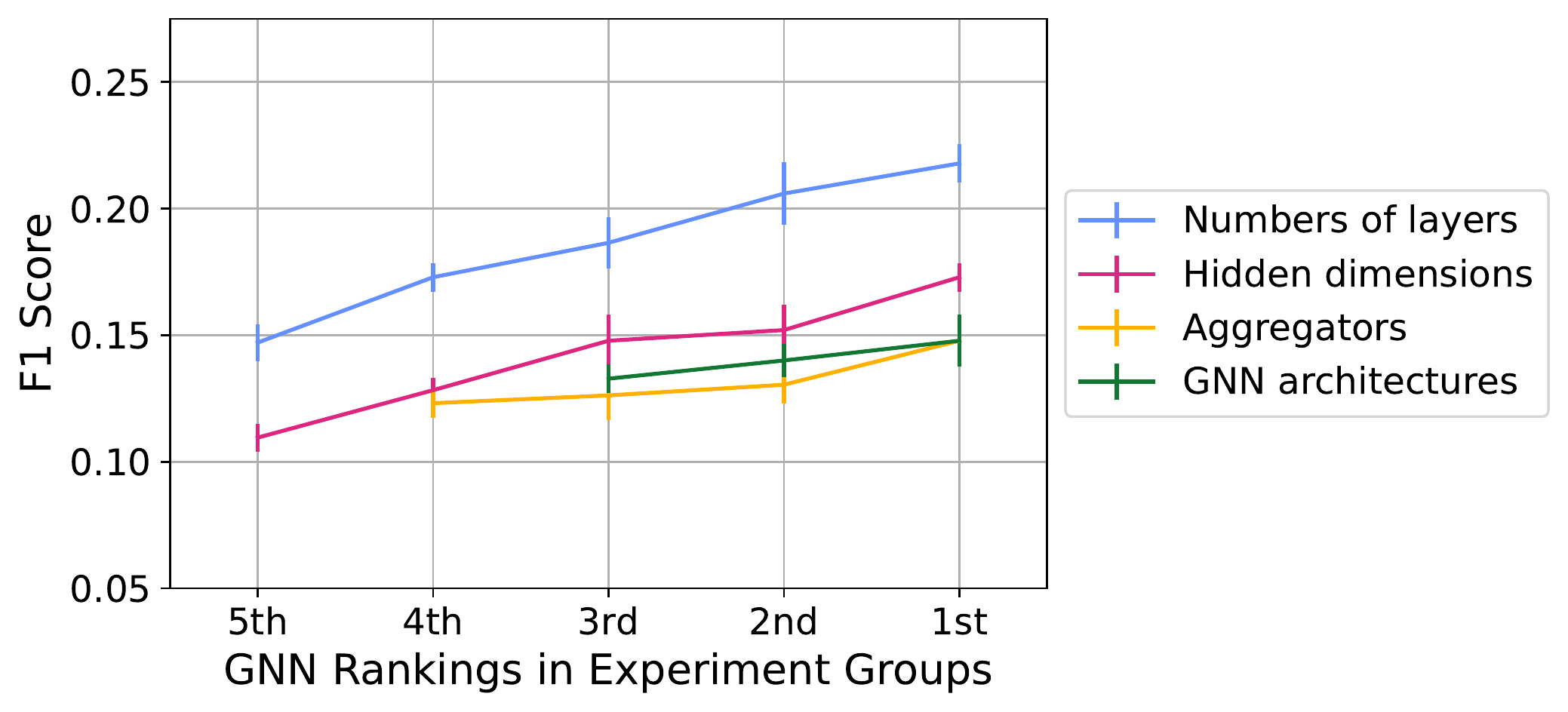}
\end{figure}

\begin{figure}[h]
    \centering
    \begin{subfigure}[t]{0.49\textwidth}
        \includegraphics[scale=0.45, valign=t]{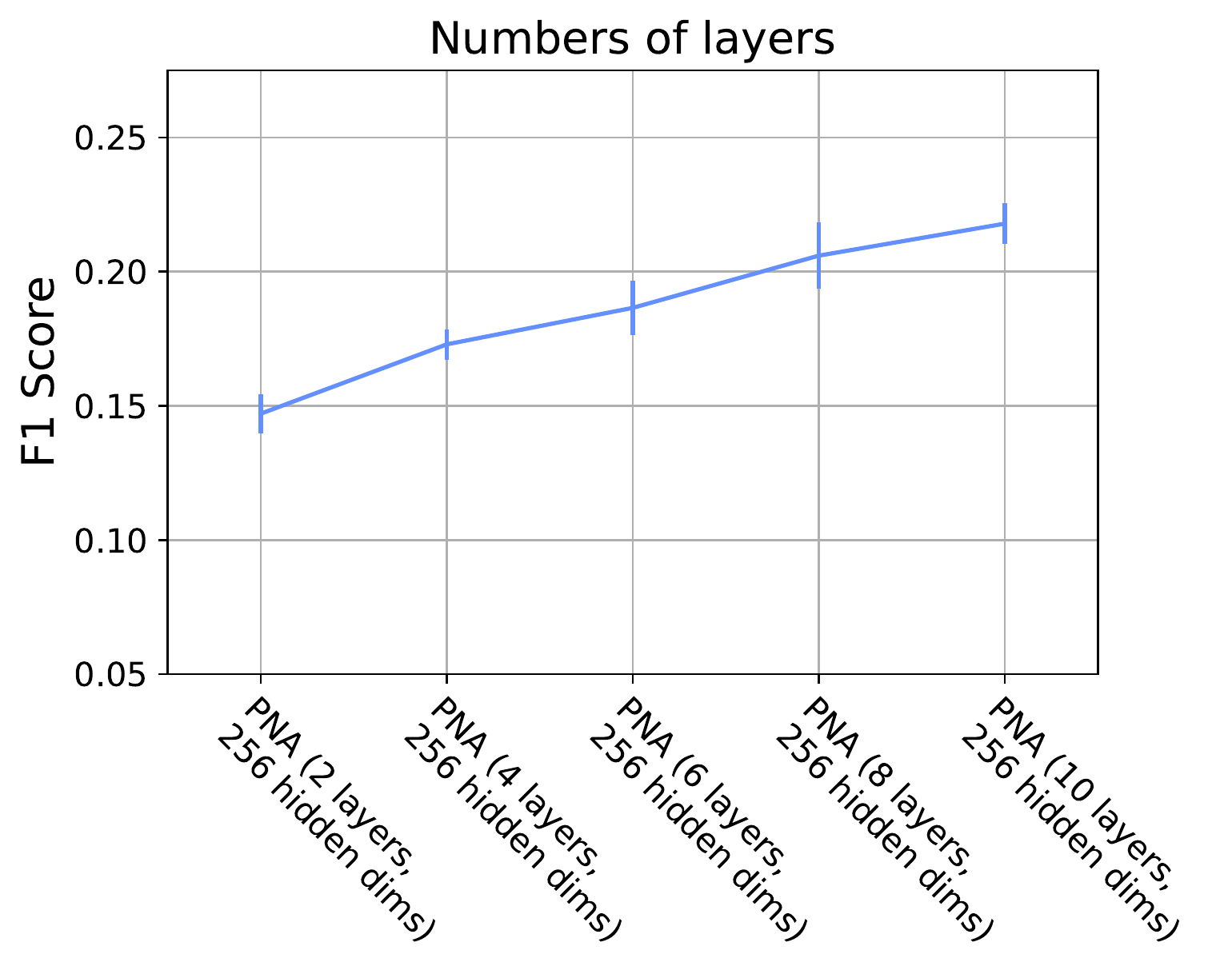}
    \end{subfigure}
    \begin{subfigure}[t]{0.49\textwidth}
        \includegraphics[scale=0.45, valign=t]{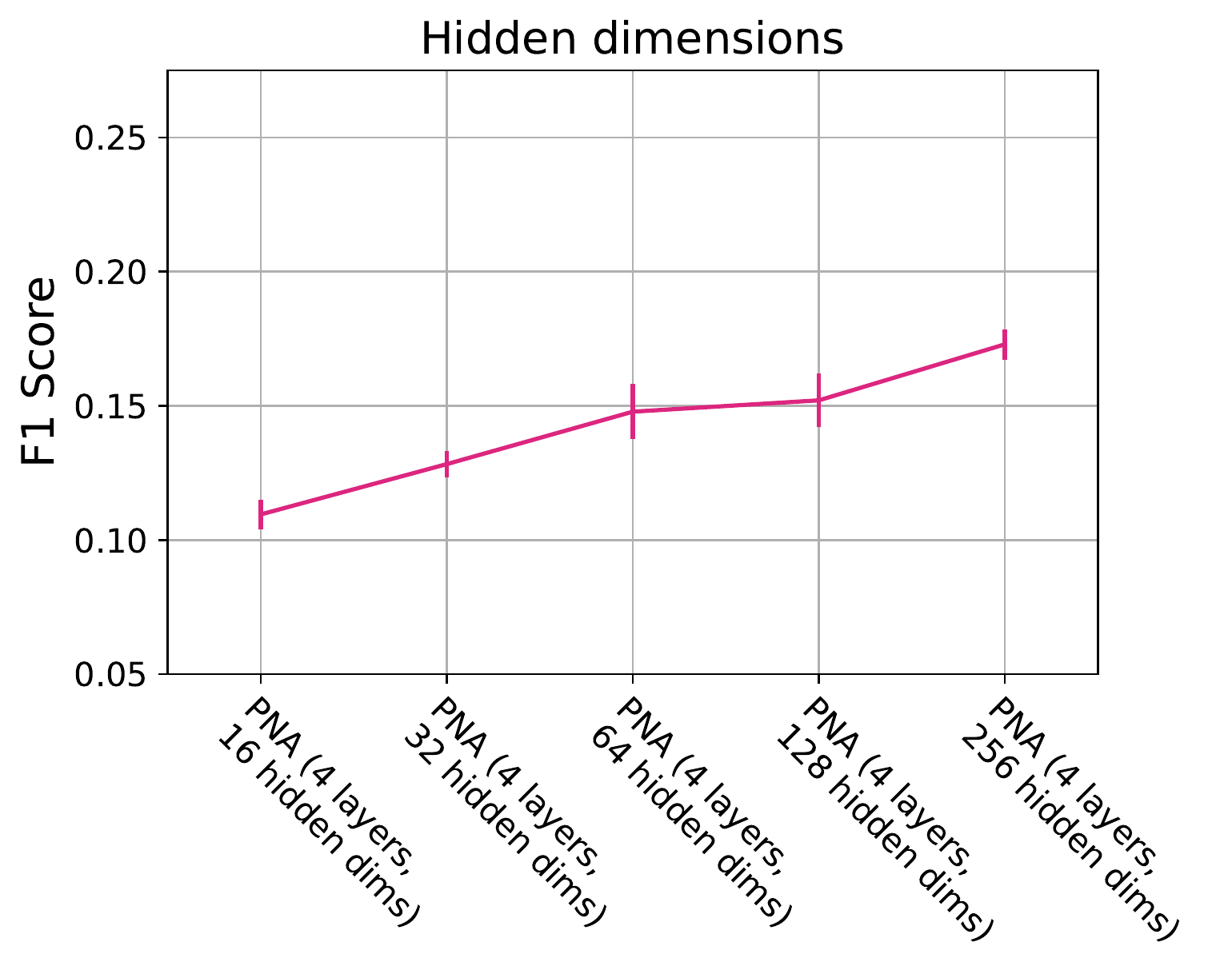}
    \end{subfigure}
    \begin{subfigure}[t]{0.49\textwidth}
        \includegraphics[scale=0.45, valign=t]{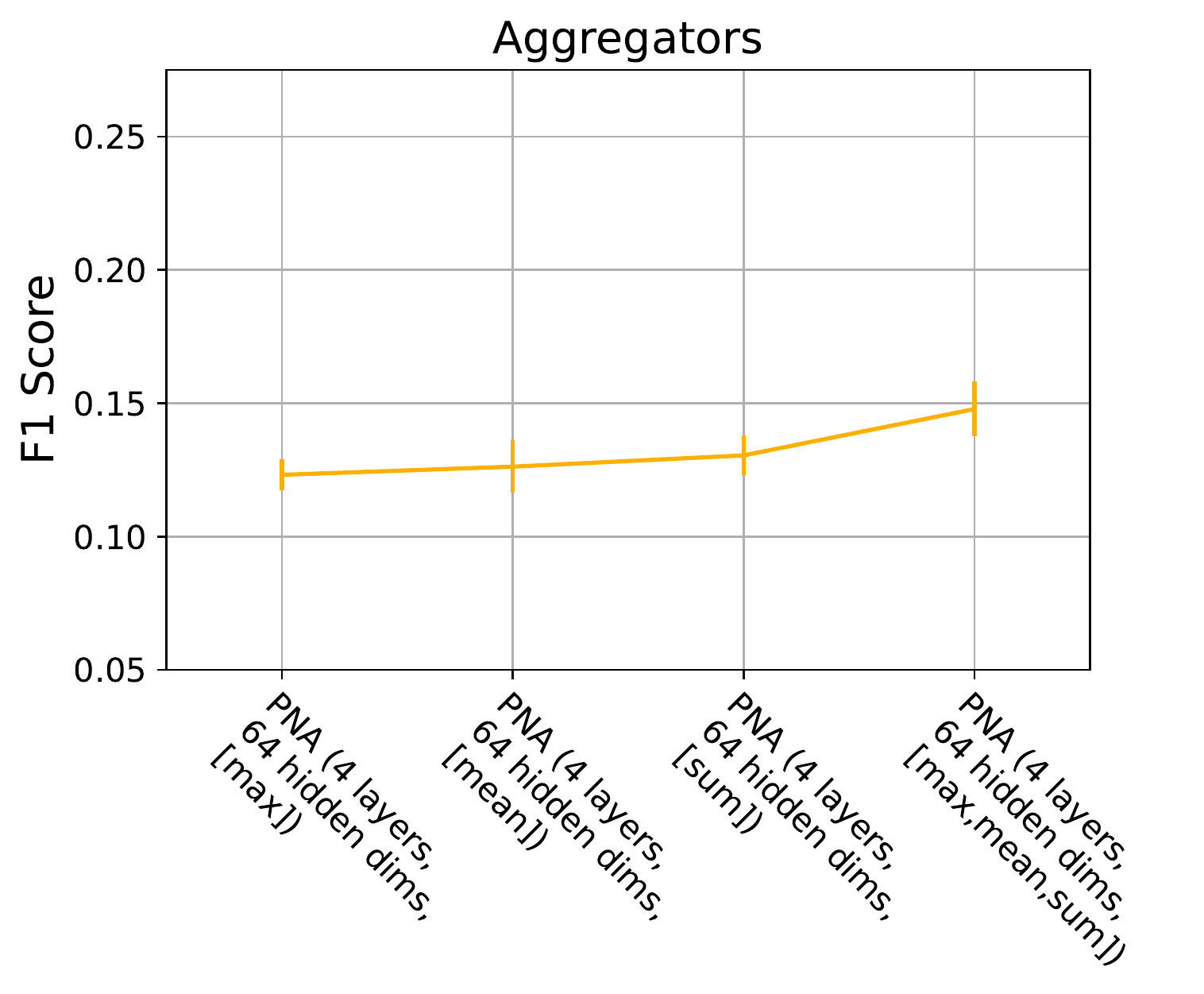}
    \end{subfigure}
    \begin{subfigure}[t]{0.49\textwidth}
        \includegraphics[scale=0.45, valign=t]{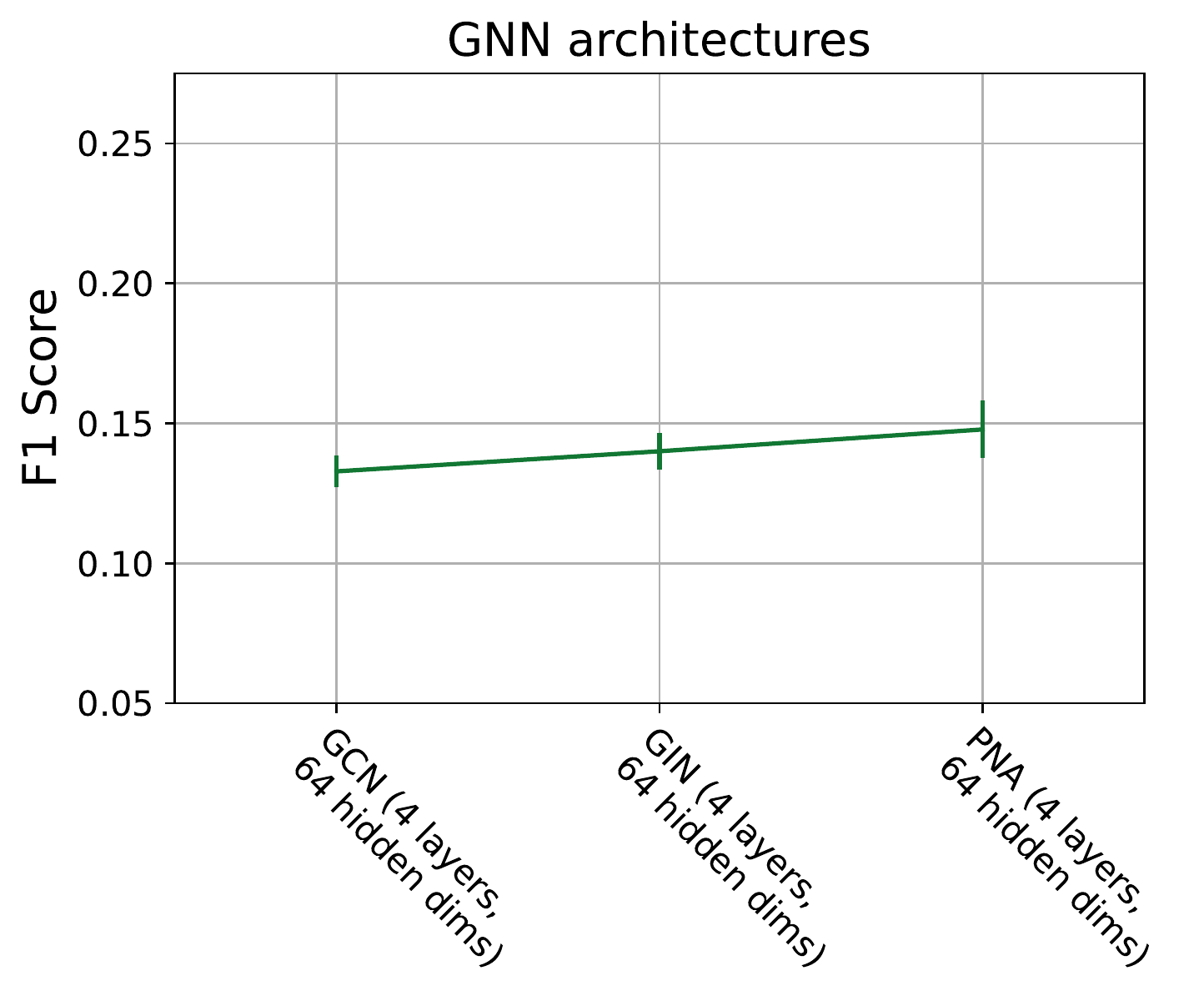}
    \end{subfigure}
    \caption{Correlation plots of GraphAC's GNN rankings with the GNNs' task performances on the \texttt{ogbg-code2} dataset, with detailed descriptions of the GNN architectures and parameters.}
    \label{fig:correlation-detailed-code2}
\end{figure}

\newpage

%% file: appendices/training-details.tex
\section{Training Details} 

\subsection{Hyperparameter Tuning} \label{sec:hyperparam-tuning}

For all hyperparameter tuning experiments, we use a 10-layer PNA with 256 hidden dimensions, and a 10-layer PNA with 128 hidden dimensions, as the pair of competing GNNs. Both PNAs use [max, mean, sum] as their aggregators, [identity, amplification, attenuation] as their scalers, and their message passing functions are parametrized by 2-layer MLPs. The output dimensionality is set to 256 for all experiments. These settings on the GNNs are only for standardizing hyperparameter tuning, and can be changed in the actual evaluation of GraphAC. Adam \cite{Kingma2015Adam} was used as the optimizer for all experiments. 

Hyperparameter tuning of GraphAC was focused on the batch size of the training data, weighting coefficients of the sums of the two triangles, and the learning rates. Since the trial experiments have shown that the Competitive Barlow Twins and VICReg covariance regularisation terms are in the same order or magnitude, and they share a similar importance in stabilizing training, the Competitive Barlow Twins terms $\mathcal{L}_{\text{CBT}_A}$, $\mathcal{L}_{\text{CBT}_B}$ and VICReg covariance regularisation terms $\mathcal{L}_\text{Cov}$ are set to share the same weights (i.e., $\alpha=\beta=1$ in Equation \eqref{eqn:competitive-bt-cov-loss}), and the value of $\lambda$ in Equation \eqref{eqn:competitive-bt-loss} adopts the hyperparameter tuning results in the original Barlow Twins paper \citep{Zbontar2021BarlowTwins}, which is $\text{5}\times\text{10}^{-\text{3}}$. In order to test whether trading-off is required between collaboration (i.e., predicting the other GNNs' graph embeddings) and competition (i.e., preventing the other GNNs from predicting the GNNs' own graph embeddings), the $\mu$ in Equation \eqref{eqn:competitive-bt-loss} is set to take values from 0.1, 0.2, 0.5, 1, 2, 5 and 10. In order to thoroughly validate the proposed framework for GraphAC and find the optimal hyperparameter settings for it, we conducted hyperparameter tuning experiments in a grid-search manner on each framework. The hyperparameter search space and the final values selected for GraphAC are specified in Table~\ref{tab:hyperparam}:

\vspace{\baselineskip}

\begin{table}[ht]
    \centering
	\caption{Hyperparameters searched for the competitive Barlow Twins framework. \textbf{Bold} values indicate the final selections.}
    \label{tab:hyperparam}
    \vskip 0.15in
	\begin{tabular}{llc}\toprule
		Dataset &Hyperparameter &Search space \\\midrule
        \multirow{3}{*}{\texttt{ogbg-molpcba}} &Weighting coefficient of the triangle ($\mu$) &[0.1, 0.2, 0.5, \textbf{1}, 2, 5, 10] \\
        &Batch size of the training data &[64, 128, 256, \textbf{512}, 1024] \\
		&Learning rate &[$\text{1}\times\text{10}^{-\text{5}}$, {\bf $\text{5}\times\text{10}^{-\text{5}}$}, $\text{2}\times\text{10}^{-\text{4}}$] \\\midrule
        \multirow{2}{*}{\texttt{ogbg-code2}} &Batch size of the training data &[64, \textbf{128}, 256, 512, 1024] \\
		&Learning rate &[{\bf $\text{1}\times\text{10}^{-\text{5}}$}, $\text{5}\times\text{10}^{-\text{5}}$, $\text{2}\times\text{10}^{-\text{4}}$] \\
		\bottomrule
	\end{tabular}
\end{table}

\vspace{\baselineskip}

The results show that the GraphAC framework can indeed distinguish the two GNNs with stable training, and can ensure that the more expressive GNN always has a lower loss. The result that $\mu=1$ is the most and only suitable weighting coefficient of the triangle also agrees to our derivation of Competitive Barlow Twins in Section~\ref{sec:competitive-bt}. Apart from that, we noted that the training of GraphAC was generally stable and is not sensitive towards the choice of hyperparameters. 

\newpage

\subsection{Training outcomes} \label{sec:training-outcomes}

\begin{figure}[h]
	\centering
	\includegraphics[width=0.67\columnwidth]{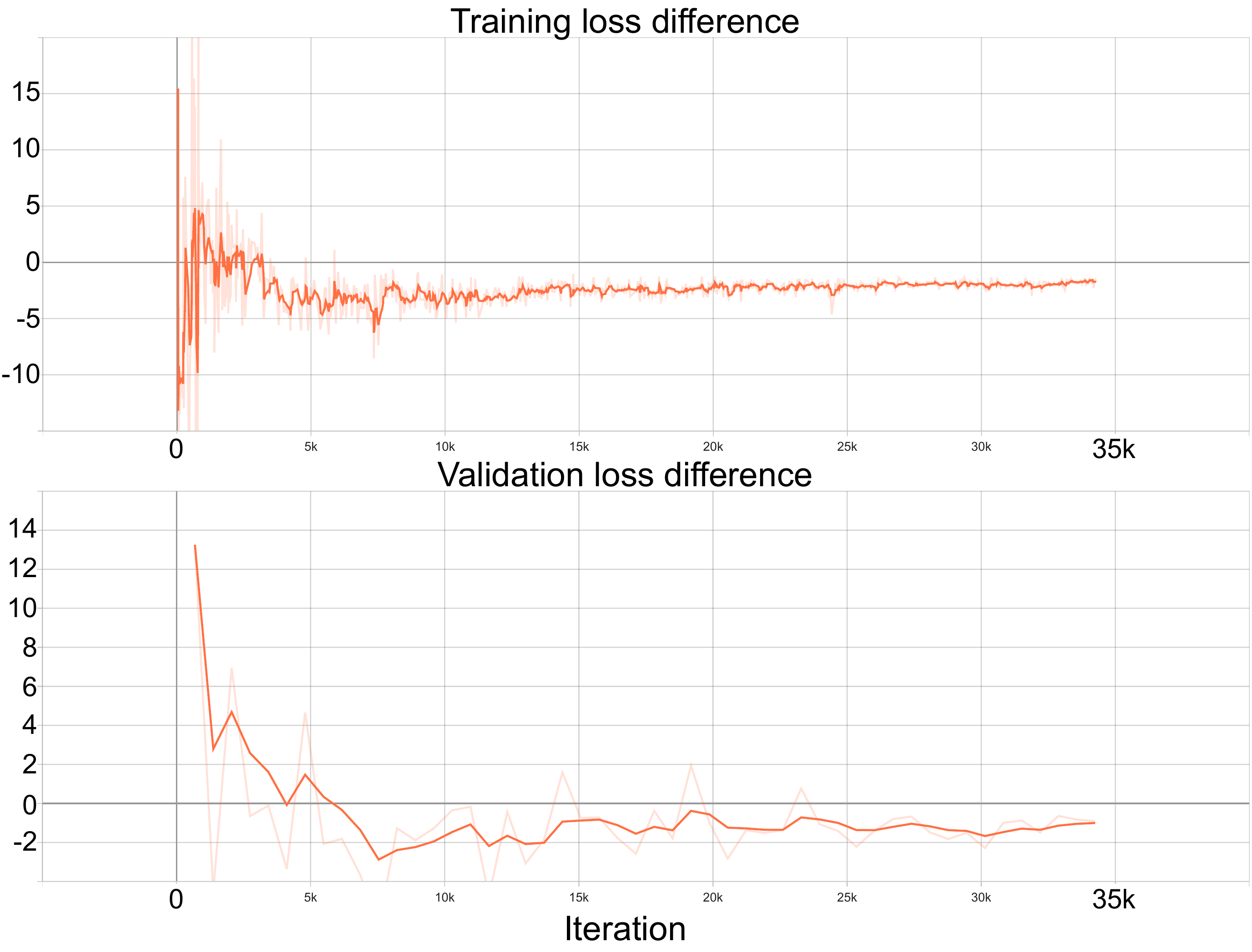}
	\caption{Example learning curves of the loss differences between the stronger and weaker GNNs under GraphAC's framework.}
	\label{fig:competitive-bt-learning-curves}
\end{figure}

\begin{figure}[h]
	\centering
	\includegraphics[width=0.67\columnwidth]{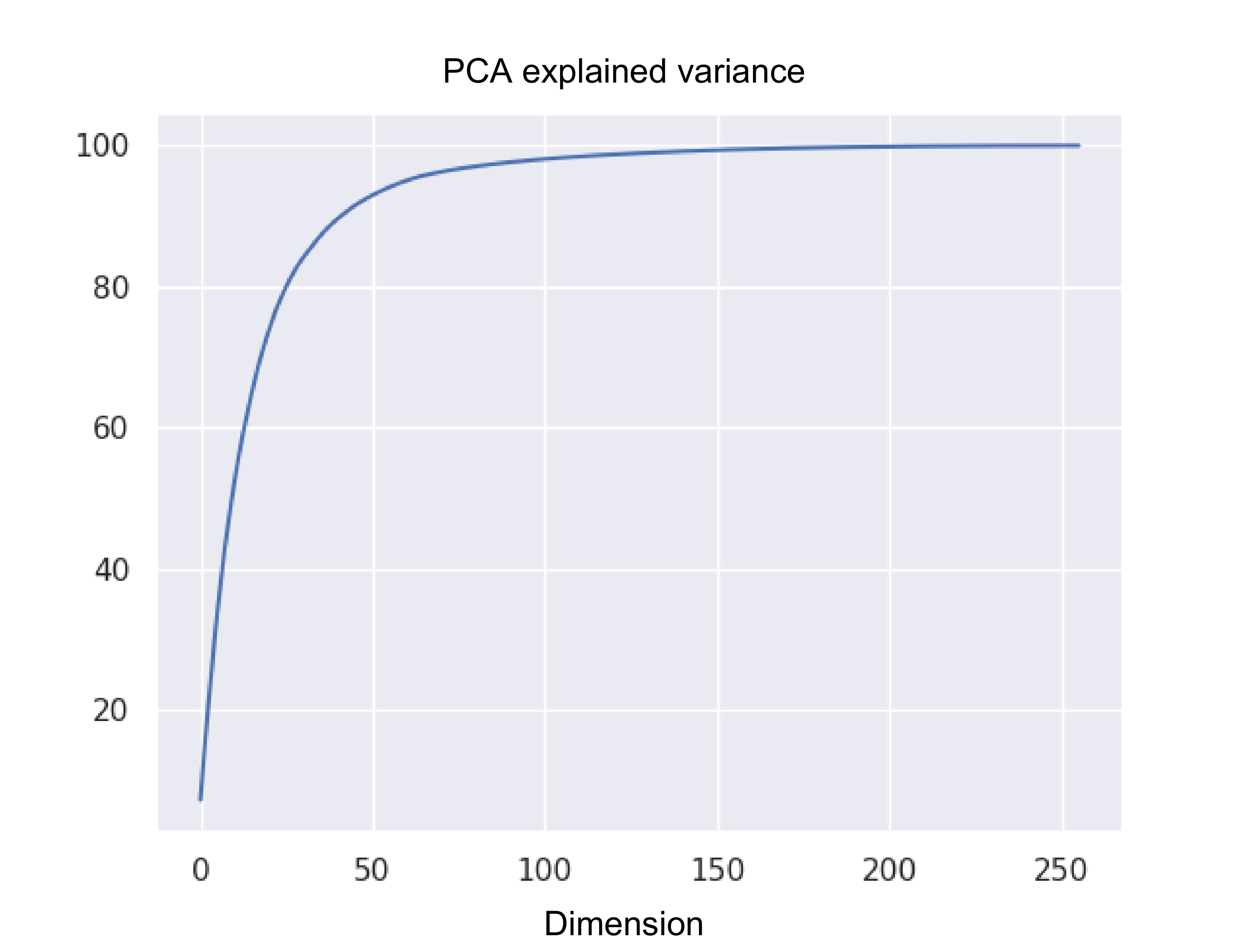}
	\caption{Example PCA explained variance of the stronger GNN's output embeddings under GraphAC's framework.}
	\label{fig:competitive-bt-pca}
\end{figure}

Figure~\ref{fig:competitive-bt-learning-curves} presents an example of learning curves of the loss difference between two GNNs with the GraphAC framework. This also indicates that the more expressive GNN can continuously achieve a lower loss in our framework. Moreover, the PCA explained variance plot in Figure~\ref{fig:competitive-bt-pca} suggests that GraphAC successfully avoids information collapse. This further confirms the claim made in Section~\ref{sec:competitive-bt} that the GNNs' output embeddings are ordered by feature importance, thereby provides a more stable training process.


%% file: graphac-iclr2023-mldd.bbl
\begin{thebibliography}{29}
\providecommand{\natexlab}[1]{#1}
\providecommand{\url}[1]{\texttt{#1}}
\expandafter\ifx\csname urlstyle\endcsname\relax
  \providecommand{\doi}[1]{doi: #1}\else
  \providecommand{\doi}{doi: \begingroup \urlstyle{rm}\Url}\fi

\bibitem[Bardes et~al.(2022)Bardes, Ponce, and LeCun]{Bardes2022VICReg}
Adrien Bardes, Jean Ponce, and Yann LeCun.
\newblock {VICR}eg: Variance-invariance-covariance regularization for
  self-supervised learning.
\newblock In \emph{10th International Conference on Learning Representations
  (ICLR 2022)}. OpenReview.net, 2022.

\bibitem[Bodnar et~al.(2021)Bodnar, Frasca, Otter, Wang, Li\`{o}, Montufar, and
  Bronstein]{Bodnar2021CWN}
Cristian Bodnar, Fabrizio Frasca, Nina Otter, Yuguang Wang, Pietro Li\`{o},
  Guido~F Montufar, and Michael Bronstein.
\newblock {Weisfeiler} and {Lehman} go cellular: {CW} networks.
\newblock In \emph{Advances in Neural Information Processing Systems (NeurIPS
  2021)}, volume~34, pp.\  2625--2640. Curran Associates, Inc., 2021.

\bibitem[Chen et~al.(2020)Chen, Kornblith, Norouzi, and Hinton]{Chen2020SimCLR}
Ting Chen, Simon Kornblith, Mohammad Norouzi, and Geoffrey Hinton.
\newblock A simple framework for contrastive learning of visual
  representations.
\newblock In \emph{Proceedings of the 37th International Conference on Machine
  Learning (ICML 2020)}, volume 119, pp.\  1597--1607. PMLR, 2020.

\bibitem[Corso et~al.(2020)Corso, Cavalleri, Beaini, Li\`{o}, and
  Veli\v{c}kovi\'{c}]{Corso2020PNA}
Gabriele Corso, Luca Cavalleri, Dominique Beaini, Pietro Li\`{o}, and Petar
  Veli\v{c}kovi\'{c}.
\newblock Principal neighbourhood aggregation for graph nets.
\newblock In \emph{Advances in Neural Information Processing Systems (NeurIPS
  2020)}, volume~33, pp.\  13260--13271. Curran Associates, Inc., 2020.

\bibitem[Dai et~al.(2021)Dai, Aggarwal, and Wang]{Dai2021NRGNN}
Enyan Dai, Charu Aggarwal, and Suhang Wang.
\newblock {NRGNN}: Learning a label noise resistant graph neural network on
  sparsely and noisily labeled graphs.
\newblock In \emph{Proceedings of the 27th ACM SIGKDD Conference on Knowledge
  Discovery and Data Mining}, pp.\  227--236. Association for Computing
  Machinery, 2021.

\bibitem[Dwivedi et~al.(2020)Dwivedi, Joshi, Laurent, Bengio, and
  Bresson]{Dwivedi2020Benchmark}
Vijay~P. Dwivedi, Chaitanya~K. Joshi, Thomas Laurent, Yoshua Bengio, and Xavier
  Bresson.
\newblock Benchmarking graph neural networks.
\newblock \emph{arXiv preprint arXiv:2003.00982v3}, 2020.

\bibitem[Dwivedi et~al.(2022)Dwivedi, Luu, Laurent, Bengio, and
  Bresson]{Dwivedi2022LSPE}
Vijay~Prakash Dwivedi, Anh~Tuan Luu, Thomas Laurent, Yoshua Bengio, and Xavier
  Bresson.
\newblock Graph neural networks with learnable structural and positional
  representations.
\newblock In \emph{10th International Conference on Learning Representations
  (ICLR 2022)}. OpenReview.net, 2022.

\bibitem[Gilmer et~al.(2017)Gilmer, Schoenholz, Riley, Vinyals, and
  Dahl]{Gilmer2017MPNN}
Justin Gilmer, Samuel~S. Schoenholz, Patrick~F. Riley, Oriol Vinyals, and
  George~E. Dahl.
\newblock Neural message passing for quantum chemistry.
\newblock In \emph{Proceedings of the 34th International Conference on Machine
  Learning (ICML 2017)}, volume~70, pp.\  1263--1272. PMLR, 2017.

\bibitem[Gutmann \& Hyv\"{a}rinen(2010)Gutmann and
  Hyv\"{a}rinen]{Gutmann2010NCE}
Michael Gutmann and Aapo Hyv\"{a}rinen.
\newblock Noise-contrastive estimation: A new estimation principle for
  unnormalized statistical models.
\newblock In \emph{Proceedings of the 13th International Conference on
  Artificial Intelligence and Statistics (AISTATS 2010)}, volume~9, pp.\
  297--304. PMLR, 2010.

\bibitem[He et~al.(2020)He, Fan, Wu, Xie, and Girshick]{He2020MoCo}
Kaiming He, Haoqi Fan, Yuxin Wu, Saining Xie, and Ross Girshick.
\newblock Momentum contrast for unsupervised visual representation learning.
\newblock In \emph{IEEE/CVF Conference on Computer Vision and Pattern
  Recognition (CVPR 2020)}, 2020.

\bibitem[Hjelm et~al.(2019)Hjelm, Fedorov, Lavoie{-}Marchildon, Grewal,
  Bachman, Trischler, and Bengio]{Hjelm2019DIM}
R.~Devon Hjelm, Alex Fedorov, Samuel Lavoie{-}Marchildon, Karan Grewal, Philip
  Bachman, Adam Trischler, and Yoshua Bengio.
\newblock Learning deep representations by mutual information estimation and
  maximization.
\newblock In \emph{7th International Conference on Learning Representations
  (ICLR 2019)}. OpenReview.net, 2019.

\bibitem[Hu et~al.(2020)Hu, Fey, Zitnik, Dong, Ren, Liu, Catasta, and
  Leskovec]{Hu2020OGB}
Weihua Hu, Matthias Fey, Marinka Zitnik, Yuxiao Dong, Hongyu Ren, Bowen Liu,
  Michele Catasta, and Jure Leskovec.
\newblock Open graph benchmark: Datasets for machine learning on graphs.
\newblock In \emph{Advances in Neural Information Processing Systems (NeurIPS
  2020)}, volume~33, pp.\  22118--22133. Curran Associates, Inc., 2020.

\bibitem[Kingma \& Ba(2015)Kingma and Ba]{Kingma2015Adam}
Diederik~P. Kingma and Jimmy Ba.
\newblock Adam: A method for stochastic optimization.
\newblock In \emph{3rd International Conference on Learning Representations
  (ICLR 2015)}, 2015.

\bibitem[Kipf \& Welling(2017)Kipf and Welling]{Kipf2017GCN}
Thomas~N. Kipf and Max Welling.
\newblock Semi-supervised classification with graph convolutional networks.
\newblock In \emph{5th International Conference on Learning Representations
  (ICLR 2017)}. OpenReview.net, 2017.

\bibitem[Kreuzer et~al.(2021)Kreuzer, Beaini, Hamilton, L\'{e}tourneau, and
  Tossou]{Kreuzer2021Rethinking}
Devin Kreuzer, Dominique Beaini, Will Hamilton, Vincent L\'{e}tourneau, and
  Prudencio Tossou.
\newblock Rethinking graph transformers with spectral attention.
\newblock In \emph{Advances in Neural Information Processing Systems (NeurIPS
  2021)}, volume~34, pp.\  21618--21629. Curran Associates, Inc., 2021.

\bibitem[NT et~al.(2019)NT, Choong, and Murata]{NT2019GNNNoisy}
Hoang NT, Jun~Jin Choong, and Tsuyoshi Murata.
\newblock Learning graph neural networks with noisy labels.
\newblock In \emph{ICLR 2019 Workshop on Learning from Limited Labeled Data},
  2019.

\bibitem[Oord et~al.(2018)Oord, Li, and Vinyals]{Oord2018InfoNCE}
Aaron van~den Oord, Yazhe Li, and Oriol Vinyals.
\newblock Representation learning with contrastive predictive coding.
\newblock \emph{arXiv preprint arXiv:1807.03748}, 2018.

\bibitem[St{\"a}rk et~al.(2021)St{\"a}rk, Beaini, Corso, Tossou, Dallago,
  G{\"u}nnemann, and Li{\`o}]{Stark20213DInfomax}
Hannes St{\"a}rk, Dominique Beaini, Gabriele Corso, Prudencio Tossou, Christian
  Dallago, Stephan G{\"u}nnemann, and Pietro Li{\`o}.
\newblock {3D Infomax} improves {GNNs} for molecular property prediction.
\newblock \emph{arXiv preprint arXiv:2110.04126}, 2021.

\bibitem[Stokes et~al.(2020)Stokes, Yang, Swanson, Jin, Cubillos-Ruiz, Donghia,
  MacNair, French, Carfrae, Bloom-Ackermann, Tran, Chiappino-Pepe, Badran,
  Andrews, Chory, Church, Brown, Jaakkola, Barzilay, and
  Collins]{Stokes2020Antibody}
Jonathan~M. Stokes, Kevin Yang, Kyle Swanson, Wengong Jin, Andres
  Cubillos-Ruiz, Nina~M. Donghia, Craig~R. MacNair, Shawn French, Lindsey~A.
  Carfrae, Zohar Bloom-Ackermann, Victoria~M. Tran, Anush Chiappino-Pepe,
  Ahmed~H. Badran, Ian~W. Andrews, Emma~J. Chory, George~M. Church, Eric~D.
  Brown, Tommi~S. Jaakkola, Regina Barzilay, and James~J. Collins.
\newblock A deep learning approach to antibiotic discovery.
\newblock \emph{Cell}, 180\penalty0 (4):\penalty0 688--702, 2020.

\bibitem[Sun et~al.(2020)Sun, Hoffmann, Verma, and Tang]{Sun2020InfoGraph}
Fan{-}Yun Sun, Jordan Hoffmann, Vikas Verma, and Jian Tang.
\newblock {InfoGraph}: Unsupervised and semi-supervised graph-level
  representation learning via mutual information maximization.
\newblock In \emph{8th International Conference on Learning Representations
  (ICLR 2020)}. OpenReview.net, 2020.

\bibitem[Veli\v{c}kovi\'{c} et~al.(2019)Veli\v{c}kovi\'{c}, Fedus, Hamilton,
  Li{\`{o}}, Bengio, and Hjelm]{Velickovic2019DGI}
Petar Veli\v{c}kovi\'{c}, William Fedus, William~L. Hamilton, Pietro Li{\`{o}},
  Yoshua Bengio, and R.~Devon Hjelm.
\newblock Deep graph infomax.
\newblock In \emph{7th International Conference on Learning Representations
  (ICLR 2019)}. OpenReview.net, 2019.

\bibitem[Weisfeiler \& Leman(1968)Weisfeiler and Leman]{Weisfeiler1968WL}
Boris Weisfeiler and Andrei Leman.
\newblock A reduction of a graph to canonical form and an algebra arising
  during this reduction.
\newblock \emph{Nauchno-Technicheskaya Informatsia}, 2\penalty0 (9):\penalty0
  12--16, 1968.
\newblock English translation available at
  \url{https://www.iti.zcu.cz/wl2018/pdf/wl_paper_translation.pdf}.

\bibitem[Xu et~al.(2019)Xu, Hu, Leskovec, and Jegelka]{Xu2019GIN}
Keyulu Xu, Weihua Hu, Jure Leskovec, and Stefanie Jegelka.
\newblock How powerful are graph neural networks?
\newblock In \emph{7th International Conference on Learning Representations
  (ICLR 2019)}. OpenReview.net, 2019.

\bibitem[Xu et~al.(2021)Xu, Wang, Ni, Guo, and Tang]{Xu2021GraphLoG}
Minghao Xu, Hang Wang, Bingbing Ni, Hongyu Guo, and Jian Tang.
\newblock Self-supervised graph-level representation learning with local and
  global structure.
\newblock In \emph{Proceedings of the 38th International Conference on Machine
  Learning (ICML 2021)}, volume 139, pp.\  11548--11558. PMLR, 2021.

\bibitem[You et~al.(2020)You, Chen, Sui, Chen, Wang, and Shen]{You2020GraphCL}
Yuning You, Tianlong Chen, Yongduo Sui, Ting Chen, Zhangyang Wang, and Yang
  Shen.
\newblock Graph contrastive learning with augmentations.
\newblock In \emph{Advances in Neural Information Processing Systems (NeurIPS
  2020)}, volume~33, pp.\  5812--5823. Curran Associates, Inc., 2020.

\bibitem[You et~al.(2021)You, Chen, Shen, and Wang]{You2021GraphCLAutomated}
Yuning You, Tianlong Chen, Yang Shen, and Zhangyang Wang.
\newblock Graph contrastive learning automated.
\newblock In \emph{Proceedings of the 38th International Conference on Machine
  Learning (ICML 2021)}, volume 139, pp.\  12121--12132. PMLR, 2021.

\bibitem[Zbontar et~al.(2021)Zbontar, Jing, Misra, LeCun, and
  Deny]{Zbontar2021BarlowTwins}
Jure Zbontar, Li~Jing, Ishan Misra, Yann LeCun, and Stephane Deny.
\newblock Barlow twins: Self-supervised learning via redundancy reduction.
\newblock In \emph{Proceedings of the 38th International Conference on Machine
  Learning (ICML 2021)}, volume 139, pp.\  12310--12320. PMLR, 2021.

\bibitem[Zhang et~al.(2017)Zhang, Bengio, Hardt, Recht, and
  Vinyals]{Zhang2017Understanding}
Chiyuan Zhang, Samy Bengio, Moritz Hardt, Benjamin Recht, and Oriol Vinyals.
\newblock Understanding deep learning requires rethinking generalization.
\newblock In \emph{5th International Conference on Learning Representations
  (ICLR 2017)}. OpenReview.net, 2017.

\bibitem[Zhang et~al.(2020)Zhang, Hu, Shi, and Wang]{Zhang2020Adversarial}
Mengmei Zhang, Linmei Hu, Chuan Shi, and Xiao Wang.
\newblock Adversarial label-flipping attack and defense for graph neural
  networks.
\newblock In \emph{2020 IEEE International Conference on Data Mining (ICDM
  2020)}, pp.\  791--800, 2020.

\end{thebibliography}
